\documentclass{article}

\usepackage{microtype}
\usepackage{graphicx}

\usepackage{booktabs} 
\usepackage{caption}
\usepackage{subcaption}
\usepackage{multirow}
\usepackage{wrapfig}

\usepackage{amsmath}
\usepackage{amssymb}
\usepackage{mathtools}
\usepackage{amsthm}
\usepackage{graphics}
\usepackage[export]{adjustbox}
\usepackage{placeins}

\usepackage[symbol]{footmisc}

\PassOptionsToPackage{numbers, compress, sort}{natbib}
\usepackage[
    pagebackref=true,
    breaklinks=true,
    colorlinks,
    urlcolor=blue,
    linkcolor=red,
    bookmarks=false]{hyperref}

\usepackage[capitalize,noabbrev]{cleveref}

\usepackage[preprint]{neurips_2024}

\usepackage[utf8]{inputenc} 
\usepackage[T1]{fontenc}    
\usepackage{url}            
\usepackage{booktabs}       
\usepackage{amsfonts}       
\usepackage{nicefrac}       
\usepackage{microtype}      
\usepackage{xcolor}         

\usepackage{caption}
\usepackage{subcaption}
\usepackage{multirow}

\usepackage[textsize=tiny]{todonotes}

\title{Learning rigid-body simulators over implicit shapes\\for large-scale scenes and vision}

\author{
Yulia Rubanova \quad Tatiana Lopez-Guevara \quad Kelsey R. Allen \quad William F. Whitney \\
\textbf{Kimberly Stachenfeld} \quad \textbf{Tobias Pfaff}  \\
\vspace{0.2cm}
\centering Google Deepmind
}

\begin{document}

\maketitle

\begin{abstract}

Simulating large scenes with many rigid objects is crucial for a variety of applications, such as robotics, engineering, film and video games. Rigid interactions are notoriously hard to model: small changes to the initial state or the simulation parameters can lead to large changes in the final state. Recently, \textit{learned} simulators based on graph networks (GNNs) were developed as an alternative to hand-designed simulators like MuJoCo \citep{todorov2012mujoco} and PyBullet \citep{Coumans2015}. They are able to accurately capture dynamics of real objects directly from real-world observations. However, current state-of-the-art learned simulators operate on meshes and scale poorly to scenes with many objects or detailed shapes. Here we present SDF-Sim, the first learned rigid-body simulator designed for scale. We use learned signed-distance functions (SDFs) to represent the object shapes and to speed up distance computation. We design the simulator to leverage SDFs and avoid the fundamental bottleneck of the previous simulators associated with collision detection.
For the first time in literature, we demonstrate that we can scale the GNN-based simulators to scenes with hundreds of objects and up to 1.1 million nodes, where mesh-based approaches run out of memory. Finally, we show that SDF-Sim can be applied to real world scenes by extracting SDFs from multi-view images.

\end{abstract}


\section{Introduction}

Simulating real-world environments, such as a bookshelf with books and decorations or a dinner table with plates and glasses, presents a significant challenge for traditional physics simulators. These simulators require precise 3D shape, location, and corresponding physical parameters of each object, making the simulation of arbitrary scenes a difficult task.

Recently, \textit{learned} simulators based on graph networks (GNNs) \cite{pfaff2021learning, allen2023fignet, fignetplus} have been introduced as a powerful alternative to traditional hand-designed simulators like MuJoCo~\citep{todorov2012mujoco} or PyBullet~\citep{Coumans2015}. Graph networks can capture a range of complex physical dynamics, learn physical properties directly from real data and generalize to new scenes. However, current GNN-based methods do not scale well to large scenes. Similarly to the traditional simulators, they rely on 3D meshes to describe object shapes. In scenes with a large number of objects or objects with detailed meshes, these scenes might contain thousands of nodes and mesh triangles, making collision detection between objects extremely costly.
In the context of graph networks, large meshes also lead to input graphs that might contain hundreds of thousands of nodes and edges, crippling runtime and memory performance.

To remedy the issue with expensive collision detection, a common approach in the classic simulation literature is to use \textit{signed-distance functions} or \textit{fields} (SDFs). SDFs implicitly represent the object shapes and allow to find the distance and the closest point on the object surface from an arbitrary location in 3D space in constant-time. Since distance queries are a main driver of the compute cost in traditional rigid body simulation, they can be significantly sped up using SDFs.
However, in practice, SDFs are frequently pre-computed from a mesh and stored as a 3D grid, which trades off the runtime efficiency for increased memory cost and limits their usefulness for large real scenes.

An orthogonal line of work started to explore SDFs \textit{learned} from a set of images to reconstruct the 3D shape from vision. Those SDFs store the shape implicitly in the MLP weights and are fast to train and query. They are less memory-hungry compared to 3D grids, making them a perfect candidate for simulation. However, despite these advantages, learned SDFs were applied for dynamics scenes only in a limited context \cite{SURFSUP,driess2021learning}.

\begin{figure}
\centering
\includegraphics[width=0.95\textwidth]{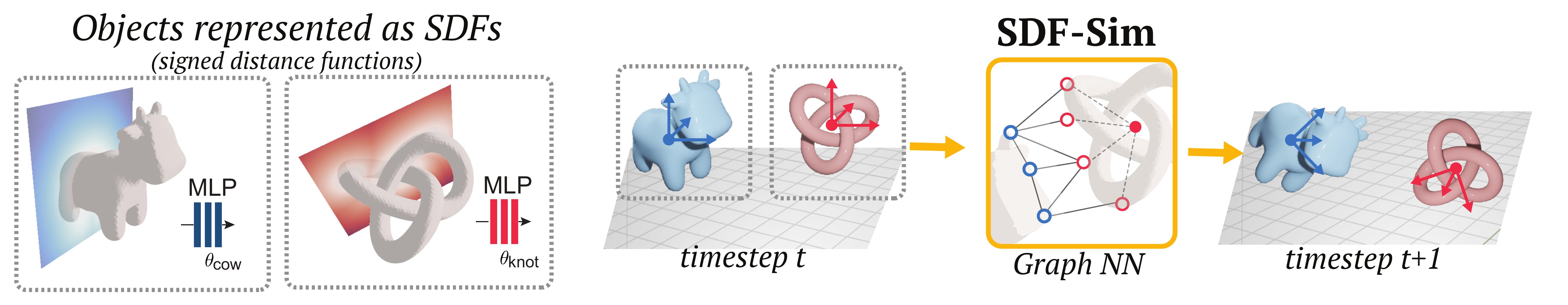} \vspace{-3pt}
\caption{Overview of SDF-Sim pipeline. SDFs parameterized by MLPs are learned for each object to implicitly represent the object shape and the distance field. A GNN-based simulator uses learned SDFs to predict object dynamics for the next simulation step.
}
\label{fig:overview}
\end{figure}

We present SDF-Sim, a learned graph-network-based simulator for rigid interactions that uses learned SDFs to represent object shapes, Figure~\ref{fig:overview}. Using a special design of the input graph, SDF-Sim allows us to substantially reduce the runtime and memory requirements compared to mesh-based learned simulators. This is the first demonstration of a learned simulator generalizable to extremely large scenes: 851k nodes in Figure~\ref{fig:large_falling_pile} and up to 1.1 million nodes in Figure~\ref{app:large_falling_pile}, orders of magnitude larger than what have been shown in any previous work on learned rigid simulation. Finally, we show that SDF-Sim can directly work with SDFs obtained from multi-view RGB images, Figure~\ref{fig:sdf_from_vision}, supporting rich 3D simulation of objects extracted from real scenes.

\newcommand{\rolloutwidth}{0.32\textwidth}
\begin{figure*}
\centering
\includegraphics[width=\rolloutwidth]{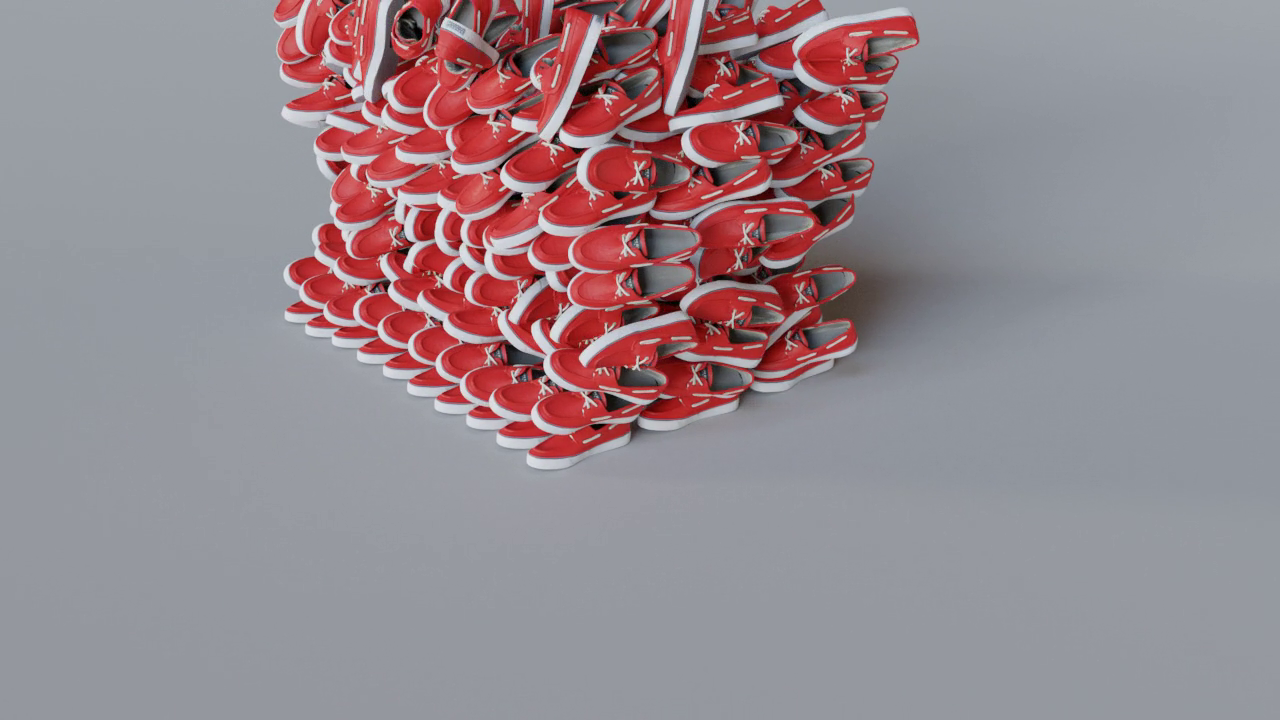}
\includegraphics[width=\rolloutwidth]{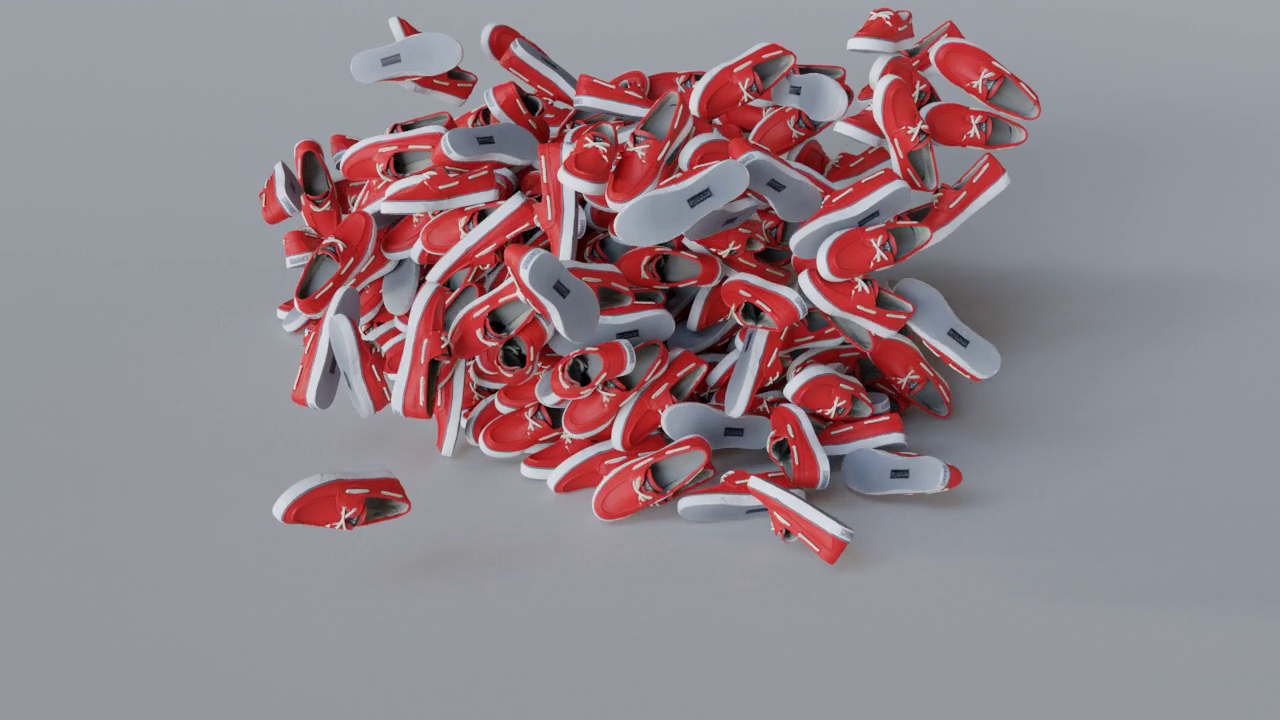}
\includegraphics[width=\rolloutwidth]{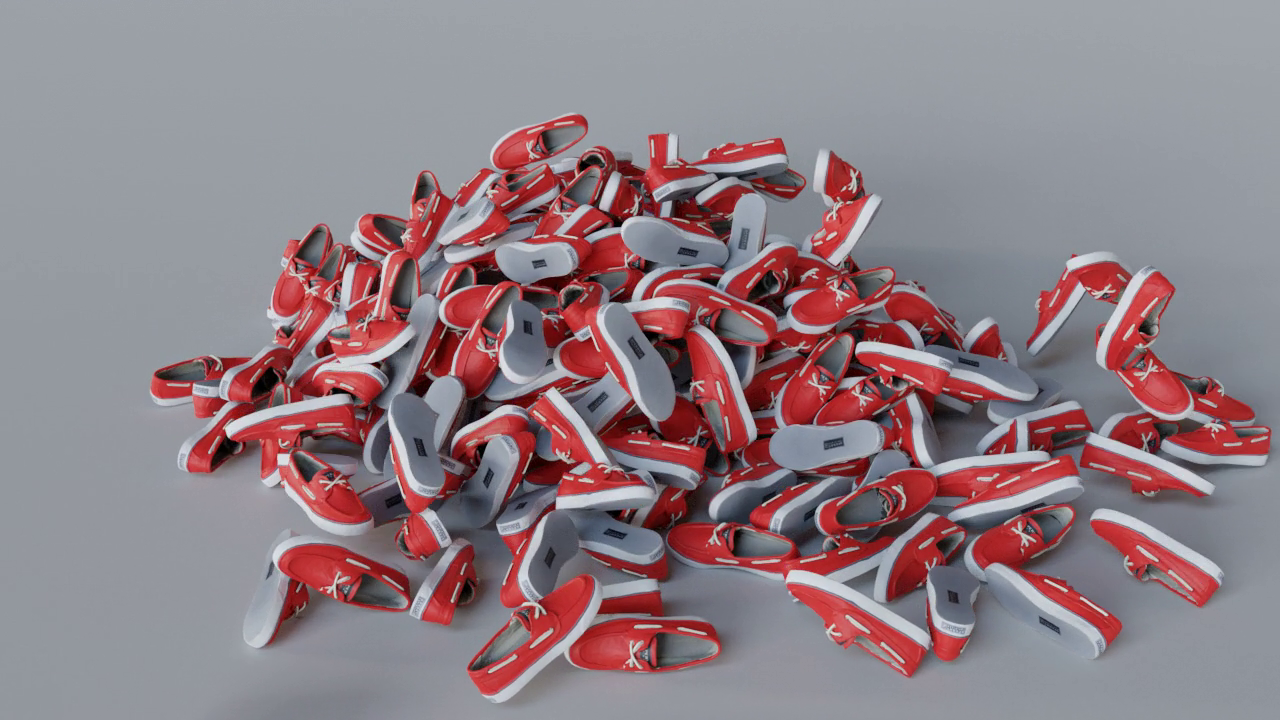}
\caption{Example of rollouts from SDF-Sim scaled to large simulations, all simulated for 200 steps. 300 shoes (object from Movi-C), with 851k nodes, falling onto the floor. See more examples with up to 1.1 million nodes in Figure~\ref{app:large_falling_pile} and simulation videos on {\small\url{ https://sites.google.com/view/sdf-sim}}.}
\label{fig:large_falling_pile}
\vspace{0.2cm}
\includegraphics[width=\textwidth]{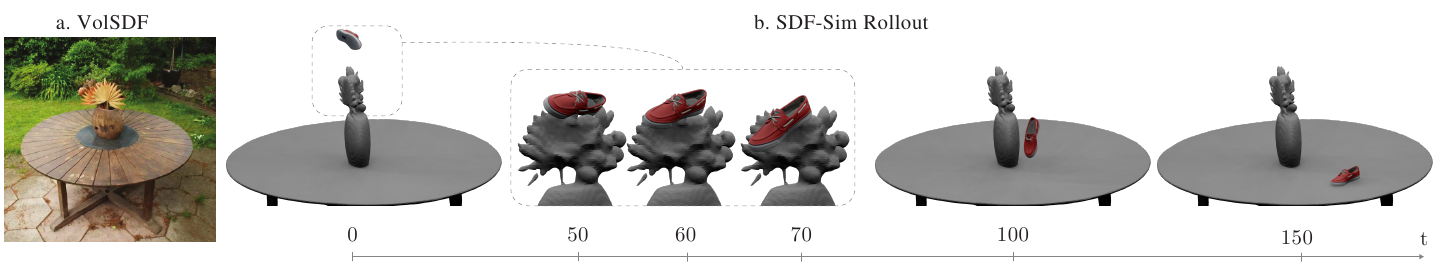}\vspace{-5pt}
\centering
\caption{Simulating assets extracted from vision. (a) We extract the SDF from the images of a real-world scene with a garden table  \citep{mipnerf360}. (b) We simulate a virtual shoe object falling onto a vase and a table using SDF-Sim. SDF-Sim is able to predict realistic dynamics, even for the collision of the shoe with the intricate shape of the vase (frames 50-70). See section~\ref{sec:sdf_vision} for details and the video on the \href{https://sites.google.com/view/sdf-sim}{website}.}
\label{fig:sdf_from_vision}
\end{figure*}

\section{Background}

\paragraph{Mesh-based simulation}
\label{sec:meshsim}

A simulation can be represented as a time series of system states $\mathcal{S}^1, \ldots, \mathcal{S}^T$. 
The goal is to learn a neural simulator that predicts the next state $\mathcal{S}^{t+1}$  given a history of previous states $\{\mathcal{S}^{t-h+1},\ldots,\mathcal{S}^t\}$.
Simulators based on graph networks (GNNs) \citep{pfaff2021learning,battaglia2018relational} encode the system state into a graph $\mathcal{G}=(\mathcal{V}, \mathcal{E})$ with nodes $\mathcal{V}$ and edges $\mathcal{E}$. For rigid body simulation, this graph can be constructed from the triangle mesh of the individual objects: mesh vertices become the graph nodes, and mesh edges act as graph edges. The object motion is computed by message passing across the nodes and edges in the graph. Within individual objects, the position, rotation and velocity of the object can be computed by message passing through the nodes and edges of that object.

\paragraph{Modelling collisions in GNN simulators}
The most challenging component of the simulation is computation of collision impulses between objects. To do so, GNN simulators introduce collision edges $\mathcal{E}^{coll}$ between nodes~\citep{sanchez2020learning} or triangles~\citep{allen2023fignet} on the mesh surface that belong to different objects and are within a predefined distance threshold. However, the amount of these edges is the main bottleneck of GNN-based simulators. Asymptotically, the number of potential collision edges grows quadratically with the number of simulated nodes, leading to prohibitive compute and memory costs.
%
Another challenge is identifying which pairs of triangles/nodes to connect with collision edges in the first place, by computing the distance to the closest point on a mesh. Typically, this procedure is implemented with tree search methods over all mesh triangles in the scene, such as  BVH~\citep{bvh}.
This computation is expensive and often relies on CPU-implementations that are difficult to accelerate or integrate into deep learning pipelines. In this work, we address both of these challenges by using SDFs.

\paragraph{Signed distance functions}
Signed-distance functions (SDFs) are widely used in computer graphics, game engines, and robotics for fast collision detection and computation of distances to an object \cite{Zhang2021DynamicNG,Spelunking_the_deep}. SDF defines a field $f(\mathbf{y}):~\mathbb{R}^3 \rightarrow \mathbb{R}$ that represents the signed distance from an arbitrary point $\mathbf{y}$ to the closest point on the surface of the object.
The sign of the SDF determines whether a point is outside (positive) or inside (negative) of the object. The zero level of the SDF $\{\mathbf{y} \in \mathbb{R}^3 | f(\mathbf{y}) = 0 \}$ implicitly represents the object surface.
SDFs permit constant-time queries of the distance to an object surface, irrespective of the number of nodes/faces in the object mesh, which is an essential component of collision detection.

\paragraph{Constructing an SDF} In the computer graphics and simulation literature, SDF is often pre-computed as a high-resolution 3D grid containing signed distances from the points on the grid to the object surface \cite{turpin2023fastgraspd,murali2023cabinet}. The grid SDF allows to speed up distance queries by trading off memory: for example, 512x512x512 grid would take $\approx$134M voxels (0.5Gb of memory) for a single object.

Learned SDFs started to gain popularity for reconstruction of water-tight 3D shapes from images. They approximate the continuous distance field $f(\mathbf{y}; \theta)$ with an MLP parameterized by $\theta$ \citep{volsdf, deepSDF, SURFSUP}. Unlike 3D-grid-SDFs, learned SDFs are not tied to a fixed grid resolution and can represent detailed shapes using a small set of parameters $\theta$.
Despite these advantages, \textit{learned} SDFs have not been sufficiently explored to speed up physical simulations. Limited available works combine learned SDFs with classic physics solvers \citep{BerticheArgila2021NeuralIS, Spelunking_the_deep}. They demonstrate that learned SDFs can massively reduce the distance query time thanks to parallelization on a GPU compared to mesh-based computation, while taking $\sim$32x less memory than traditional 3D-voxel-grid SDFs.

\paragraph{Computing closest points}
For any point in 3D space $\mathbf{y}$ an SDF $f_{\theta}$ allows us to easily compute the closest point $\mathbf{y}^*$ on the object surface that it represents, as:
\begin{equation}
\setlength{\abovedisplayskip}{3pt}
\setlength{\belowdisplayskip}{5pt}
\mathbf{y}^* = \mathbf{y} - f_{\theta}(\mathbf{y}) \nabla f_{\theta}(\mathbf{y}),
\label{eq:sdf_proj}
\end{equation}
where, by definition, $f_{\theta}(\mathbf{y})$ is the distance between $\mathbf{y}$ and $\mathbf{y}^*$, and the gradient $\nabla f_{\theta}(\mathbf{y})$ points in the opposite direction of the shortest path from $\mathbf{y}$ to the surface of the object and is unit-norm.
 If $f_{\theta}(\mathbf{y})$ is parameterized as an MLP, this calculation requires only one forward and one backward pass of the network. This provides an efficient way to calculate the closest points for collision resolution in the simulation.
In this work, for the first time in the literature, we use learned SDFs to accelerate distance computation in SOTA graph-network simulators.

\section{SDF-Sim}
\label{sec:sdfsim}

We introduce SDF-Sim, a graph-network-based simulator for rigid objects that uses learned SDFs to represent object shapes and to perform fast and memory-efficient distance computation between the objects. By leveraging SDF properties, we propose a new way to construct the input graph for the graph network, allowing us to use a smaller graph size and to get an order-of-magnitude reduction in runtime and memory on large simulations.

\subsection{Training SDF functions per object}

We represent SDF $f_{\theta}(\mathbf{y})$ as an MLP that takes in a 3D point $\mathbf{y} \in \mathbb{R}^3$ and outputs a scalar SDF value. Learned SDFs are pre-trained separately for each object and remain fixed throughout the simulation.

\textbf{\textit{Meshes}} To compare to the existing mesh-based baselines, we apply SDF-Sim on benchmarks where meshes are available and train the learned SDF $f_{\theta}(\mathbf{y})$ from a mesh.
To train an SDF, we sample points in 3D space and compute the ground-truth signed distances from these points to the mesh surfaces using a classic BVH \cite{bvh} algorithm. Finally, we train an MLP $f_{\theta}(\mathbf{y})$ to fit these distances with supervision. See more details in section \ref{sec_app:sdf_training}.
We use the same architecture and the model size for each object shape of 8 MLP layers with 128 units each, unless otherwise stated.

\textbf{\textit{Vision}} We use VolSDF \citep{volsdf} to distill an SDF from a set of images representing a 360-degree view of the outdoor garden scene first described in \citep{mipnerf360} with camera poses estimated via COLMAP \citep{colmap}. See visualisation in Figure \ref{fig:sdf_from_vision} and section \ref{app:sdf_vision} for details.

\subsection{Learned simulator}


\begin{figure}
\vspace{-0.6cm}
\includegraphics[width=1.0\textwidth]{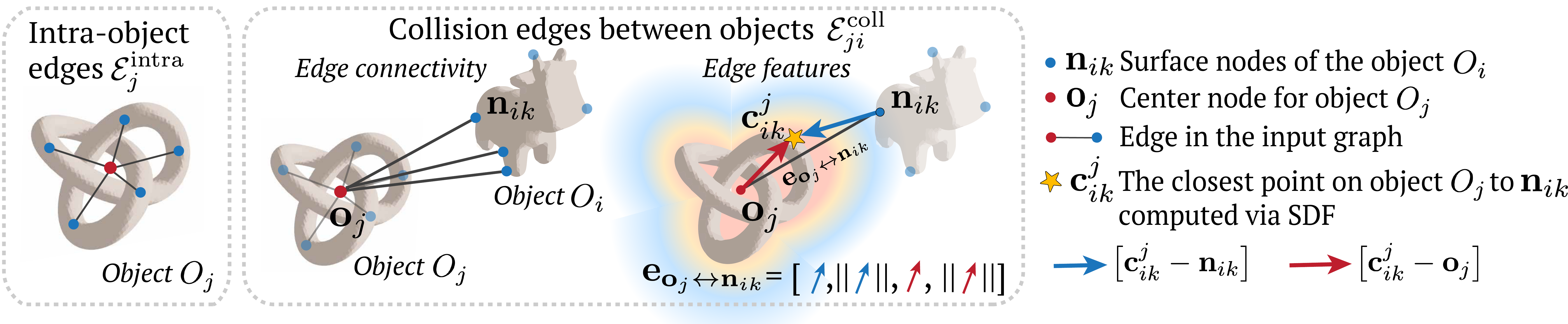}
\caption{Construction of graph edges in SDF-Sim.}
\label{fig:sdf_edge_overview}
\vspace{-0.3cm}
\end{figure}

\paragraph{Object representation}
We represent the shape of the object $O_i$ in the reference pose (centered at zero) as a learned SDF function $f_{\theta_i}$. At a simulation step $t$, a rigid transformation $\mathcal{T}_i^t = (\mathbf{p}_i^t, \mathbf{R}_i^t)$ transforms the object from reference pose to the current pose in the simulation. Here, $\mathbf{p}_i^t$ is an object translation, corresponding to the object's center of mass at timestep $t$; $\mathbf{R}_i^t$ is a rotation. The task of the learned simulator is predicting the next-step transformation $\mathcal{T}_i^{t+1}$ for each object. In the following text, we omit the time index $t$ for brevity.

To represent $I$ objects in the input graph, we 
introduce \emph{object} nodes $\mathcal{V}_O = \{\mathbf{o}_i | i=1..I\}$, located at position $\mathbf{p}_i$, and a set of \emph{surface} nodes $\mathcal{V}_S = \{ \mathbf{n}_{ik} \in O_i | i=1..I, k=1..K_i\}$ on the surface of the objects, where the number of nodes $K_i$ may differ for each object. These surface nodes $\{\mathbf{n}_{ik}\}$ move with their corresponding object according to the transformation $\mathcal{T}_i^t$. With a slight abuse of notation, we will refer to $\{\mathbf{n}_{ik}\}$ both as the node entities and their 3D position in the simulation space.

\paragraph{Nodes and edges within the object} To construct the graph connectivity \textit{within} an object, we follow an established line of work on learned simulators \cite{fignetplus,allen2023fignet,pfaff2021learning}. We connect surface nodes $\{\mathbf{n}_{ik} \}$ to their corresponding object node $\mathbf{o}_i$ using edges $\mathcal{E}^{\mathrm{intra}}_i = \{ \mathbf{e}_{\mathbf{o}_i \leftrightarrow \mathbf{n}_{ik}} | \mathbf{n}_{ik} \in O_i \}$. Thus, all the information about object motion, e.g., impulses from collision events, is propagated between the nodes via the object node $\mathbf{o}_i$. As shown by \cite{fignetplus} we can omit the surface edges between the nodes $\{\mathbf{n}_{ik}\}$ without loss of accuracy.

In graph networks, nodes and edges are associated with feature vectors that can hold the information about the motion and the relation between the nodes. We follow the approach of \cite{fignetplus,allen2023fignet,pfaff2021learning} to construct a set of nodes and edge features that are necessary for simulation: To construct the node features for surface nodes $\mathbf{n}_{ik}$, we compute the finite-difference velocity estimates in the simulation space using a history of the latest three timesteps  $\mathbf{v}_{ik} = (\mathbf{n}^t_{ik} - \mathbf{n}^{t-1}_{ik},\mathbf{n}^{t-1}_{ik} - \mathbf{n}^{t-2}_{ik})$. We set node features to be $[\mathbf{v}_{ik}, ||\mathbf{v}_{ik}||, \phi_i]$ where $\phi_i$ are the constant object parameters: mass, friction and restitution. We use the same procedure for the object nodes $\{\mathbf{o}_i\}$ using their positions $\mathbf{p}_i$.
For intra-object edges $\mathcal{E}^{\mathrm{intra}}_i$, we use displacement vector between the surface node position and the object center as the edge feature $\mathbf{e}_{\mathbf{o}_i \leftrightarrow \mathbf{n}_{ik}} = [\mathbf{o}_i - \mathbf{n}_{ik}, ||\mathbf{o}_{i} - \mathbf{n}_{ik}||]$.
 
\paragraph{SDF-based inter-object edges.}

Here we introduce a new way to construct collision edges between the objects by leveraging their SDF representations (Figure \ref{fig:sdf_edge_overview}). We design these edges such that they contain sufficient information to detect collisions, while their number remains linear in the number of nodes. This is unlike quadratic number of collision edges in mesh-based simulators.

Consider two objects $O_i$ and $O_j$. For a node $\mathbf{n}_{ik}$ on $O_i$, we directly query the SDF $f_{\theta_j}$ of object $O_j$ to get the distance $d_{ik}^j$ from $\mathbf{n}_{ik}$ to the closest point on $O_j$. We note that unlike mesh-based approaches, this is a single test, and we do not need to calculate the distance from $\mathbf{n}_{ik}$ to \emph{each} node/triangle on $O_j$. Then, if this distance $d_{ik}^j$ is within a predefined distance threshold $\mathcal{D}$, we connect the surface node $\mathbf{n}_{ik}$ directly to the opposing \emph{object} node $\mathbf{o}_j$ and refer to this edge as $\mathbf{e}_{\mathbf{o}_j \leftrightarrow \mathbf{n}_{ik}}$.
Thus, we define the set of inter-object edges as $\mathcal{E}^{\mathrm{coll}}_{ji} = \{ \mathbf{e}_{\mathbf{o}_j \leftrightarrow \mathbf{n}_{ik}} | \forall \mathbf{n}_{ik} \in O_i : f_j(\mathbf{n}_{ik}) \leq \mathcal{D} \}$. This approach is different from mesh-based simulators \cite{fignetplus,allen2023fignet,pfaff2021learning} that connect pairs of nodes or triangles on the two surfaces.
 
We construct the features for collision edges $\mathbf{e}_{\mathbf{o}_j \leftrightarrow \mathbf{n}_{ik}}$ such that they contain information about potential points of collision.
First, we compute the closest point $\mathbf{c}^j_{ik}$ from $\mathbf{n}_{ik} \in O_i$ to the surface of object $O_j$. To do so, we transform the position of $\mathbf{n}_{ik}$ into the reference space of $O_j$ using $\mathcal{T}_j^{-1}(\mathbf{n}_{ik})$. We call an SDF function to get the distance from the node $\mathbf{n}_{ik}$ to the closest point on $O_j$ as $d_{ik}^j = f_{\theta_j}(\mathcal{T}_j^{-1}(\mathbf{n}_{ik}))$. The closest point on the surface of $O_j$ is then computed similarly to Eq.~\ref{eq:sdf_proj}:
\begin{equation}
\setlength{\abovedisplayskip}{1pt}
\setlength{\belowdisplayskip}{1pt}
\mathbf{c}_{ik}^j = \mathbf{n}_{ik} - d^j_{ik} \mathcal{T}_j \Bigl(\nabla f_{\theta_j}\left( \mathcal{T}_j^{-1}(\mathbf{n}_{ik}) \right) \Bigr)
\end{equation}
Note that $\mathbf{c}_{ik}^j$ lies on the surface defined by $f_{\theta_j}$, but does not have to coincide with any surface node and is not part of node set $\mathcal{V}$. Thus, SDFs allow us to test contact between two objects at higher fidelity, without relying on the density of the surface nodes.

Finally, we construct the features for the collision edges as follows: $\mathbf{e}_{\mathbf{o}_j \leftrightarrow \mathbf{n}_k} = [ \mathbf{c}^j_{ik} - \mathbf{n}_{ik}, \mathbf{c}^j_{ik} - \mathbf{o}_j, ||\mathbf{c}^j_{ik} - \mathbf{n}_{ik}||,  ||\mathbf{c}^j_{ik} - \mathbf{o}_j||]$. These features provide information about the relative position of the closest point (a potential collision point) within the object $O_j$ as well as its relative location to the node $\mathbf{n}_{ik}$ on the opposing object. This information is sufficient for the neural network to resolve collisions. Through message-passing over such input graph, the object node $\mathbf{o}_j$  can collect the information from all the nodes that are within a collision radius relative to the object $O_j$. The model also has access to the velocity and rotation history for both objects through node features, so it is able to infer how fast the node $\mathbf{n}_{ik}$ is changing its position w.r.t. $O_j$.

With such construction, SDF-Sim requires significantly fewer edges than mesh-based methods, because a single collision edge can test a node against an entire object surface. Asymptotically, SDF-Sim has $\mathcal{O}(I\cdot K)$ edges in the worst case, with the number of objects $I$ being magnitudes smaller than the total number of surface nodes $K = \sum_i K_i$, instead of $\mathcal{O}(K^2)$ for mesh-based simulators. As we demonstrate below, this choice of input graph and simulator unlocks the ability to scale to very large scenes that has not been previously shown in the literature.

\paragraph{Graph network simulator}
We encode the input graph using MLPs for each node and edge type. We process the graph using 10 message passing steps as in \cite{pfaff2021learning,allen2023fignet}. We decode the processed surface node features into an acceleration
estimate $\mathbf{a}_{ik}$ for each surface node $\mathbf{n}_{ik}$, and compute a per-node position update $\hat{\mathbf{n}}^{t+1}_{ik}$ using Euler integration. 
Finally, next-step rigid transformations $\{\hat{\mathcal{T}}^{t+1}_i\}$ are computed from $\hat{\mathbf{n}}^{t+1}_{ik}$ using shape matching~\citep{muller2005meshless}, following \cite{allen2023fignet}.
The simulator is trained on a single-step prediction task using a per-node L2 loss on the acceleration estimate $\mathbf{a}_{ik}$.

\section{Results}

\begin{wrapfigure}{r}{0.5\textwidth}
\vspace{-0.7cm}
\centering
\includegraphics[width=0.5\textwidth]{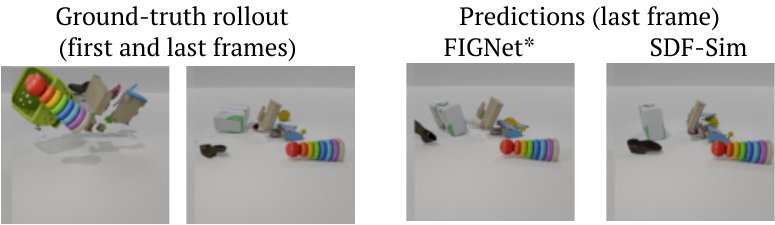}
\caption{Comparison of the last frames of rollouts predicted on Movi-C. See more frames in Figure~\ref{fig:rollouts} and on the \href{https://sites.google.com/view/sdf-sim}{website}.}
\label{fig:rollout_examples}
\vspace{-0.3cm}
\end{wrapfigure}

We start by evaluating the accuracy and efficiency of SDF-Sim on small-scale simulation benchmarks, where evaluation of the baselines is feasible as well.
Subsequently, we demonstrate the scaling properties of the models on scenes with an increasing number of objects. We show that SDF-Sim produces realistic rollouts of extremely large scenes with up to 1.1M nodes, which was not possible with the previous generation of learned simulators. Finally, we present ablations for how the quality of the learned SDFs affects the simulation.
All simulation videos are available on the website 
{\small\url{https://sites.google.com/view/sdf-sim}}.

\paragraph{Kubric datasets}

We evaluate SDF-Sim on the benchmark Kubric datasets \cite{greff2021kubric} (shared with Apache 2.0 license), consisting of simulated trajectories of rigid objects thrown towards the center of the scene. We evaluate on two difficulty tiers: Movi-B and Movi-C. Movi-B simulations contain eleven synthetic shapes (e.g., cube, torus, cow) with up to 1142 nodes each.
Movi-C comprises 930 scanned real-world household objects, including hollow, flat, or otherwise non-trivial surfaces with thousands of triangles. Both Movi B/C contain only 3-10 objects per simulation. We provide more details about mesh sizes in Movi-B/C in supplement
\ref{app:mesh_sizes}.

Note that we perform many of our quantitative comparisons on small-scale Movi-B/C datasets because these domains are small enough for us to run also baseline methods to quantify efficiency and accuracy. On Movi-C we compare only to FIGNet*, because other baselines run out of memory during training due to the large number of collision edges, as reported by \citet{allen2023fignet}.

\paragraph{Baseline models}
 
 \begin{figure*}
\includegraphics[width=\textwidth]{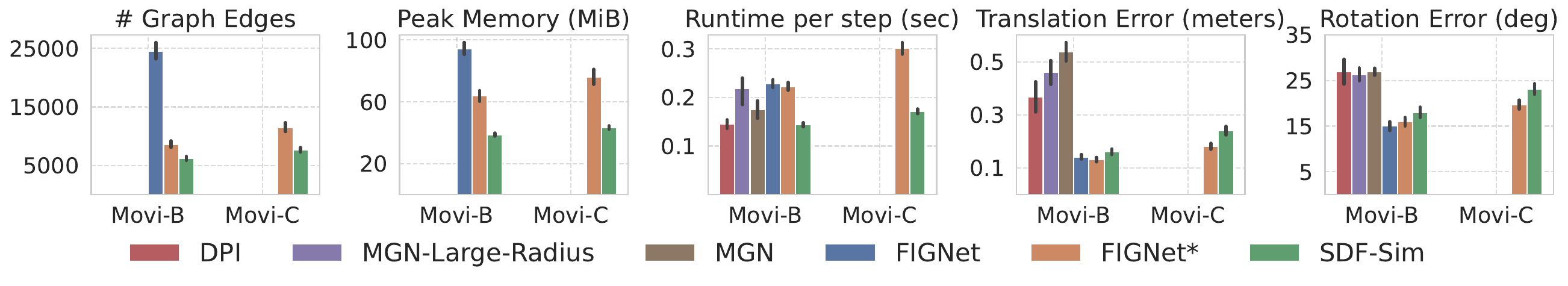}
\caption{Accuracy, memory and runtime comparisons between the SDF-Sim model and the mesh-based baselines on the Movi-B/C benchmarks. On Movi-C, most baselines except FIGNet* run out of memory and are not shown. As ``Peak Memory'' we report the peak memory used by the model per single step of the simulation. DPI, MGN-Large-Radius and MGN results were reported by \cite{allen2023fignet}. See Tables \ref{app:accuracy_comparison_movi_b} and \ref{app:accuracy_comparison_movi_c} for the exact numbers.}
\label{fig:accuracy_comparison}
\vspace{-0.3cm}
\end{figure*}

We compare SDF-Sim to the existing state-of-the-art learned simulators for rigid-body interactions: FIGNet \cite{allen2023fignet} and FIGNet* \cite{fignetplus}. Both methods are based on graph networks. Unlike SDF-Sim, they operate directly over object meshes and use a special type of graph edges between mesh triangles. Their runtime and memory costs grow with the number and size of the object meshes used for the simulation.
FIGNet* is a memory-efficient version of FIGNet that omits edges between the surface nodes of each object.
In their original publication, FIGNet was demonstrated to scale up to 10 objects with a few thousands nodes each \cite{allen2023fignet}, while FIGNet* was tested up to a larger table scene with 40k triangles, but with only one moving object \cite{fignetplus}.

We additionally include previously reported results from \citet{allen2023fignet} on Movi-B for the following models: DPI \citep{li2019propagation} that represents the objects as a set of disjoint particles; as well as MeshGraphNets (MGN) and MGN-Large-Radius \citep{pfaff2021learning}  that use inter-object edges between mesh surface nodes. 
As reported by \citet{allen2023fignet}, DPI, MGN and MGN-Large-Radius baselines suffer from a prohibitively large size of the inputs graph and can only run on small Movi-B simulations.

\subsection{Baseline comparison on small-scale datasets}
\label{sec:kubric_results}

First, we evaluate  SDF-Sim on Movi-B/C datasets with up to 10 objects. As shown in Figure \ref{fig:accuracy_comparison}, our SDF-Sim uses substantially sparser graphs to represent the scenes, with 28\% fewer graph edges compared to FIGNet* on Movi-B and 33\% fewer on Movi-C. This translates into reduction of peak memory required to execute one step of the simulation; 39\% reduction on Movi-B and 42.8\% on Movi-C. The reduction in the number of edges in SDF-Sim is enabled by object-to-node collision edges that scale linearly with the number of nodes, as opposed to quadratic number of face-face collision edges in FIGNet and FIGNet* (in the worst case, see section~\ref{sec:meshsim}). Sparser graphs in SDF-Sim also lead to lower average runtime per step of a rollout, by 36\% in Movi-B and 43\% in Movi-C, because the graph network performs edge updates for/over fewer edges even though the number of nodes in the graph remains the same. Figure~\ref{fig:runtime_comparison}(a) shows that the runtime of SDF-Sim is consistently lower than that of FIGNet* for varying sizes of the input graph. 
In the next section we show how these efficiency improvements translate into order-or-magnitude  gains on large-scale simulations.


In terms of simulation accuracy, measured as object translation and rotation RMSE errors, SDF-Sim has substantially lower errors than previous baselines DPI, MGN and MGN-Large-Radius on Movi-B. However, in comparison to SOTA models FIGNet and FIGNet*, SDF-Sim has a slightly higher error. Note that the errors of SDF-Sim are already very low: the translation error of SDF-Sim is 0.24 meters, which is only 4.9\%  of the average object travel distance of 4.92 meters in Movi-C dataset (see tables Tables~\ref{app:accuracy_comparison_movi_b} and \ref{app:accuracy_comparison_movi_c} for exact numbers). In rigid-body systems, even small discrepancies in predicted positions/rotations can lead to a drastically different object trajectory after a collision, and we consider 4.9\% deviation to be a good result. Finally, in the next section we will demonstrate that \textit{on large scenes} SDF-Sim is actually \textit{more accurate} in comparison to FIGNet*.


\subsection{SDF-Sim scales to scenes with up to 512 objects}
\label{sec:scaling_spheres}

\begin{figure*}
\centering
\includegraphics[width=\textwidth]{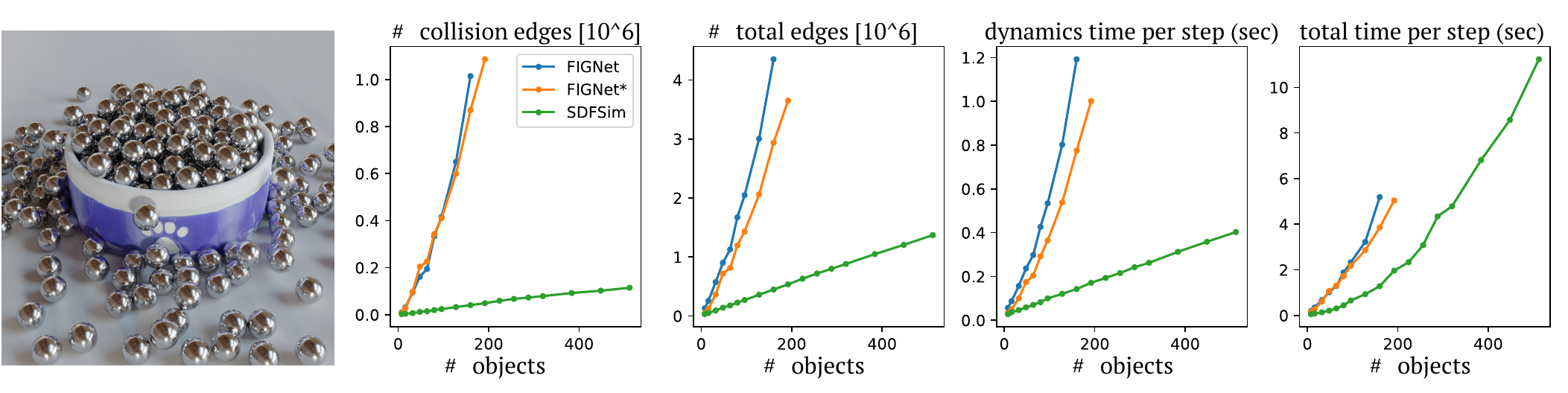}
\caption{Large-scale simulation of \textbf{Spheres-in-Bowl}, simulated for 200 timesteps. Left: the final step of SDF-Sim rollout on the scene with 512 spheres. Right: number of edges and runtime w.r.t. the number of spheres in the simulation (max 512). In complex simulations with lots of contacts, FIGNet and FIGNet* generate an excessive number of collision edges, quickly exceeding GPU memory (end of the orange and blue lines). SDF-Sim generates an order-of-magnitude fewer collision edges, and can easily simulate scenes with 100s of objects without running out of memory.}
\label{fig:large_sim_bowl}
\vspace{-0.4cm}
\end{figure*}

\begin{wrapfigure}{r}{0.6\textwidth}
\centering
\vspace{-0.5cm}
 \includegraphics[trim={10 12 15 10},clip,width=0.595\textwidth]{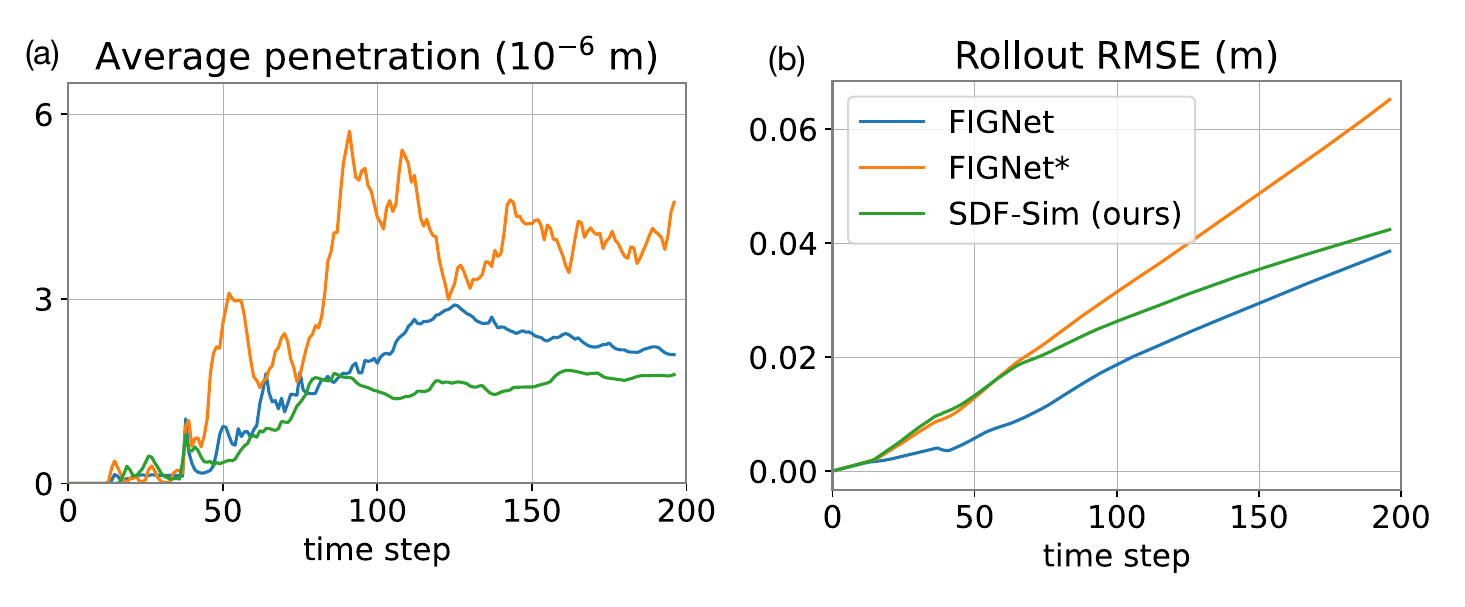}
\caption{Accuracy metrics w.r.t. simulation time step for \textbf{Spheres-in-Bowl} simulation shown in Figure~\ref{fig:large_sim_bowl}. (a) Penetration metrics. (b) Rollout RMSE. Both metrics are averaged over simulation runs with up to 140 spheres, the maximum for which all baselines could be run.}
\label{fig:sphere_in_bowl_metrics}
\vspace{-0.1cm}
\end{wrapfigure}

To study the scaling properties of the model, we created \textbf{Spheres-in-Bowl}: a set of simulations with a variable number of spheres, ranging from 1 to 512, being dropped into a bowl (Figure~\ref{fig:large_sim_bowl}). We create the ground-truth for these scenes using PyBullet, the same simulator as used for Kubric Movi-B/C. We evaluate a set of learned simulators:  SDF-Sim and FIGNet* models trained on Movi-C, as well as FIGNet trained on Movi-B. Note that the largest version, with 512 spheres, has over 50 times more objects than in Movi-B/C datasets used for training.

As shown in Figure~\ref{fig:large_sim_bowl}, SDF-Sim has an order-of-magnitude fewer collision edges, compared to FIGNet* and FIGNet.
This difference is substantial, as the number of edges dominates the memory cost. FIGNet and FIGNet* run out of memory for simulations with >140 and >160 objects, respectively, whereas SDF-Sim can simulate the entire set of 512 interacting objects using the same Nvidia V100 GPU. In terms of total runtime, SDF-Sim is up to 5 times faster than FIGNet*, even including the time for querying learned SDFs. To the best of our knowledge, this is the first demonstration of a learned simulator successfully scaling to scenes with hundreds of objects and thousands of collisions, despite being trained only on ten objects per scene.

Next, we evaluate the accuracy of these simulations.
SDF-Sim has the lowest penetration distance throughout the simulation, as shown in Figure~\ref{fig:sphere_in_bowl_metrics}(a). 
%
%
Figure~\ref{fig:sphere_in_bowl_metrics}(b) shows the rollout errors across different steps of the simulation.
Notably, SDF-Sim generalizes to these larger scenes better than FIGNet* and has lower error, despite slightly lower accuracy on smaller datasets (section \ref{sec:kubric_results}). FIGNet has the lowest error out of learned models, likely because it was trained on Movi-B, which contains a sphere object, while Movi-C dataset, used for training Fignet* and SDF-Sim, does not. 
Despite not having seen a sphere object in training, SDF-Sim performs very well on both penetration and rollout metrics, indicating that the SDF representation is not only efficient, but also shows good generalization over geometry.

\subsection{Scaling to extra-large scenes with up to 1.1 million nodes}
\label{sec:scaling_pile}
Next, we provide a further qualitative demonstration that SDF-Sim can scale to extra-large scenes and produce realistic rollouts. We design three scenes with falling stacks of shoes, metal knots, and mixed objects taken from the Movi-C dataset, shown in Figures \ref{fig:large_falling_pile} and \ref{app:large_falling_pile}. Such contact-rich simulations are challenging even for analytical simulators and are classic test examples in the computer graphics literature. These simulations consist of: 300 shoes with 851k nodes, 270 knots with 384k nodes, and 380 various Movi-C objects with  1.1M nodes, respectively. All of these scenes are orders of magnitude larger than those in Movi-C dataset used for SDF-Sim training.

Figures \ref{fig:large_falling_pile} and \ref{app:large_falling_pile} show that SDF-Sim can scale to these massive scenes without running out of memory and produce qualitatively realistic rollouts of 200 steps; see \href{https://sites.google.com/view/sdf-sim}{videos}. Note that we cannot run any other learned baseline on these scenes, as due to the large number of potential contacts, these models produce a vast number of collision edges and exceed the GPU memory.

\subsection{Simulating real-world scenes from vision}
\label{sec:sdf_vision}
\begin{figure}
\centering
\includegraphics[width=\textwidth]{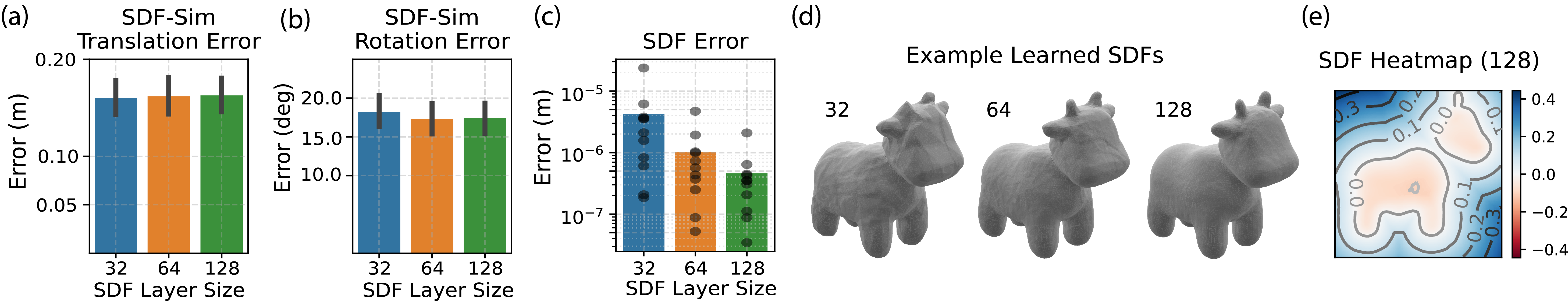}
\caption{Ablation on learned SDF model sizes: 8 layer MLPs with layer sizes of 32, 64, or 128 hidden units. (a, b) Translation and rotation error for SDF-Sim trained on Movi-B with different SDF sizes. (c) Mean squared error of the predicted SDF estimates near the surface. (d) Visualisations of the cow shape from Movi-B with different SDF sizes.
(e) A cross-section of the learned SDF field for the Movi-B cow shape.}
\vspace{5pt}
\label{fig:sdf_mlpsize}
\end{figure}

Simulating large real-world scenes is the primary area where the ability to scale to large scenes is crucial, e.g., in virtual reality applications. Here we present a proof-of-concept that SDF-Sim can successfully handle real scenes, despite being trained only on small simulated assets. We take a Garden scene \cite{mipnerf360}, represented by a sequence of RGB images and use VolSDF \citep{volsdf} to distill a 3D SDF representation of the table and the vase from this scene. SDF-Sim allows to cleanly integrate this 3D representation into a simulation, where it would interact with another object. Finally, we add a shoe object from Movi-C to this scene. Figure~\ref{fig:sdf_from_vision} shows a rollout from the SDF-Sim model of a shoe falling on top of the vase and the table SDF extracted by VolSDF. With its over 80k nodes, this mesh exceeds the limits of mesh-based methods, like FIGNet, that runs out of memory. In contrast, SDF-Sim can capture the nuanced interactions between the objects, which is particularly evident in frames 50 to 70; see the video on the  \href{https://sites.google.com/view/sdf-sim}{website}.

\subsection{Ablation of learned SDF quality}
\label{sec:learned_sdf_ablations}
We investigate the quality of the estimated SDF values and their impact on simulator performance. We train SDF using MLPs with 32, 64 and 128 units per layer, and 8 layers total. Figure~\ref{fig:sdf_mlpsize}(c) shows that larger SDF models help to improve the quality of estimated SDFs. Overall, the learned SDF models of all three layer sizes can reproduce the shape, although the reconstruction from a smallest 32-unit model is more coarse (Figure~\ref{fig:sdf_mlpsize}(d), see Figure \ref{app:sdf_ablations_all_sdfs} for more examples). However, Figure~\ref{fig:sdf_mlpsize}(a,b) shows that SDF-Sim accuracy is similar across different SDF MLP sizes, suggesting that the learned simulator can learn to compensate for noise in the estimated SDFs.  We provide additional metrics on SDF accuracy, SDF gradients and SDF projected closest points in section \ref{app:sdf_quality}. We also provide the comparison of memory footprint of SDF versus storing meshes in section \ref{sec:sdf_mesh_memory}.

\section{Related Work}

\paragraph{Dynamics over learned SDFs}

Overall, using implicit shapes, e.g., learned SDFs, for simulation has been explored only in limited settings.
DiffSDFSim \cite{DiffSDFSim} and DANO \cite{dano} use SDFs in combination with an analytic contact model on a scene with a single object. \citet{BerticheArgila2021NeuralIS} and \citet{Dynamic_mesh_aware_nerf} model the interaction of a learned SDF with a cloth via an analytic simulator. In the space of learned simulators, FIGNet* \cite{fignetplus} uses a GNN-based model on a 3D scene reconstructed via NeRF and converted into a mesh. \citet{driess2022learning} learn latent dynamics over NeRFs. \citet{vpd} convert a NeRF 3D scene into a set of particles and use them in a GNN simulator. Unlike these pipelines, SDF-Sim avoids the costly simulation of large meshes or sets of particles and operates directly over the SDFs. \citet{SURFSUP} use learned SDFs to represent large rigid scenes for the sake of efficiency and use the SDF values directly in the GNN simulator. However, this model is designed to use only a \textit{single} SDF per scene. In contrast, we focus on more challenging interactions of many rigid objects, that is much more prevalent in real applications.

\paragraph{Learning SDFs from vision}

Learned SDFs started to gain popularity for reconstructing an 3D surfaces from a sequence of images.  Unlike NeRFs \cite{mipnerf360}, SDFs are more likely to represent closed, watertight, smooth surfaces. \citet{volsdf, wang2021neus, Yu2022MonoSDF} train an MLP to represent an SDF of a single object. Other works train a single generative model for many SDF shapes, allowing to amortise the training cost, generate new shapes and perform shape completion from partial inputs
\cite{cheng2023sdfusion,chou2022diffusionsdf,Shim_2023_CVPR,deepSDF}. Recent works also offer ways to use SDFs for deformable objects \cite{Wi2022VIRDOVI}, articulated objects \cite{mu2021learning}, or to edit SDFs \citep{3D_Neural_Sculpting}. 

One can also reconstruct SDFs from other modalities, e.g., \cite{wen2023bundlesdf, mu2021learning} demonstrate how to reconstruct a new unseen object as an SDF using an RGB-D video from a single camera view. SDFs can be recovered for large room-scale scenes \cite{Azinovic_2022_CVPR} or even in real time as the camera moves through the space \cite{Ortiz:etal:iSDF2022}, or from noisy point clouds, e.g., from Lidar \cite{BaoruiNoise2NoiseMapping,Neural_Pull}. We note that although we can leverage the SDFs extracted by these methods, these works pursue a different goal and focus only on accurate reconstruction of the 3D surface and do not aim to use these SDFs for simulation.


\vspace{-0.1cm}
\section{Discussion and Limitations}

We introduce a rigid-body simulator, SDF-Sim, that uses learned SDFs to represent the object shapes and employs a graph network to learn dynamics over these objects. We re-design the construction of the input graph for the simulator to make use of SDFs and reduce the number of inter-object edges from quadratic to linear, unlocking the ability to scale to large scenes. We provide detailed study on both small-scale benchmarks and large scenes with up to 50x more objects than seen in the training.
Additionally, we investigate the impact of SDF quality on the simulation results. Our work demonstrates, for the first time in the learning simulation literature, that we can use these simulators on scenes with hundreds of objects and up to 1.1 million nodes and produce realistic rollouts, potentially unlocking applications to virtual reality, film, robotics, and more.

\textbf{Limitations} One limitation of SDF-Sim is that it requires training an SDF MLP for every new shape we want to include in a simulation. A promising future direction is to train one model on a dataset of many shapes, amortising the cost of training SDFs, similarly to \cite{deepSDF,Shim_2023_CVPR}. We also note that SDF-Sim has slightly higher error on small-scale datasets (section~\ref{sec:kubric_results}), although SDF-Sim is closer to the ground-truth on large scenes than FIGNet* (section~\ref{sec:scaling_spheres}).  We believe it is justifiable to trade a slight increase in model error in exchange for the ability to run on larger scenes than previously possible. We believe that SDF-Sim is therefore best suited for the applications that favour scaling over physical accuracy, such as animation and film. Finally, in this paper we only focus on rigid-rigid interactions, but SDFs can be extended to incorporate deformable or articulated objects \cite{Wi2022VIRDOVI,mu2021learning}, fluids \cite{SURFSUP}, and even a mix of SDFs, meshes, and particles. 

\textbf{Broader impact} We see our work as a step in making applications that require simulation, such as augmented reality or animation, more accessible to everyday consumers. As such, our work allows us to directly use 3D assets filmed on a mobile phone, and simulate large scenes using a commodity hardware with a single GPU, without requiring a specialized skill of creating meshes appropriate for simulation. Overall, we see SDF-Sim as an important step towards learning physically realistic dynamics models from real-world data, and for enabling physical reasoning over complex, large-scale physical scenes.

\clearpage

\bibliographystyle{plainnat}
\bibliography{references}

\newpage
\appendix
\onecolumn

\section{Additional large-scale simulation examples}

\begin{figure*}[hb!]
\centering
\includegraphics[width=\rolloutwidth]{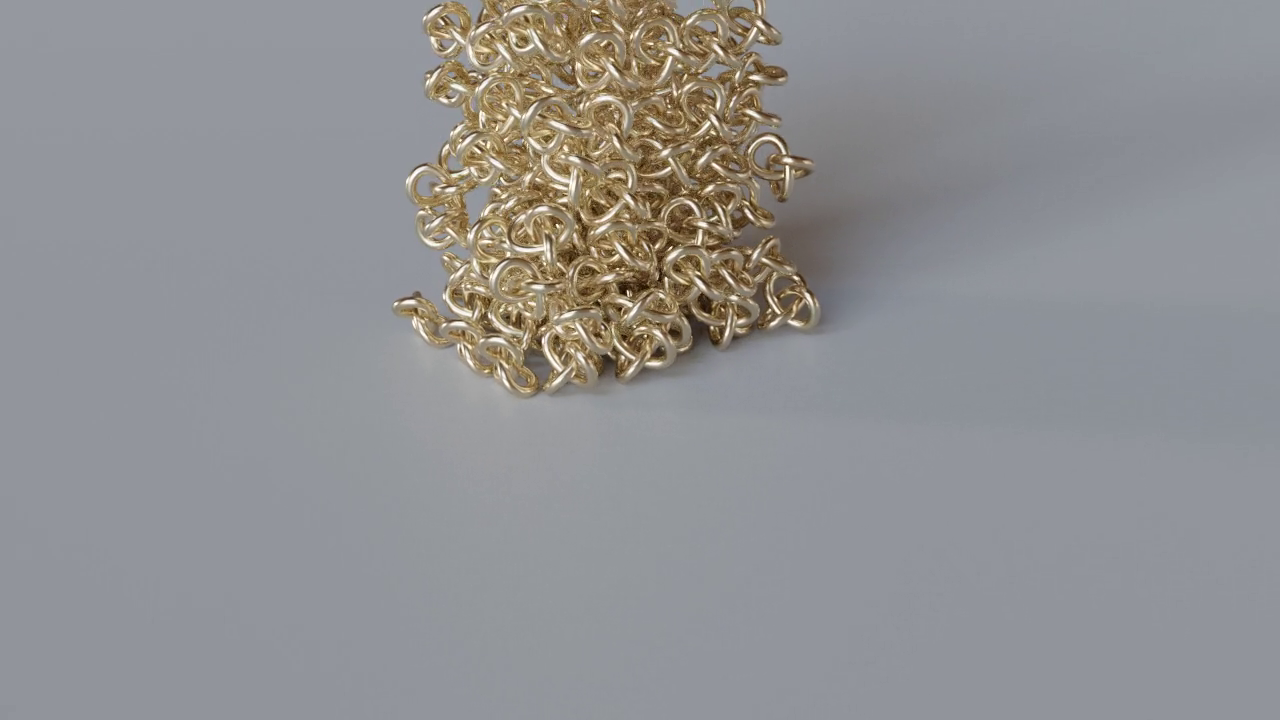}
\includegraphics[width=\rolloutwidth]{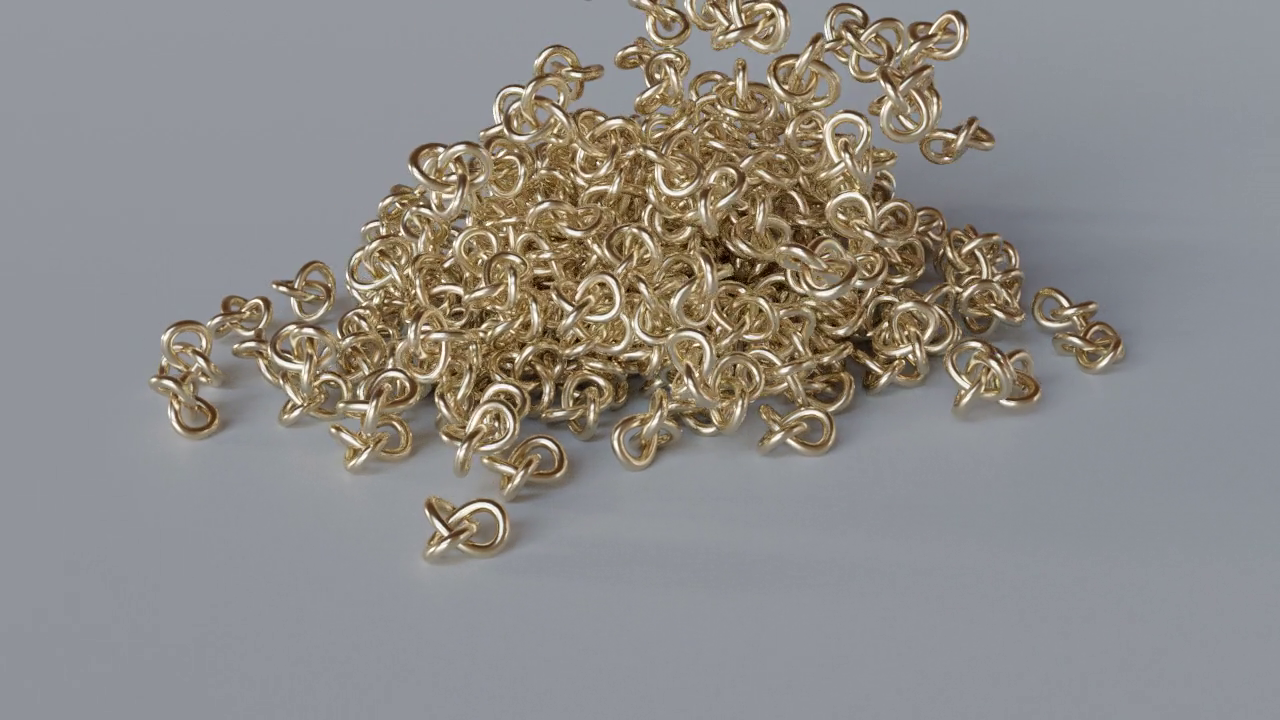}
\includegraphics[width=\rolloutwidth]{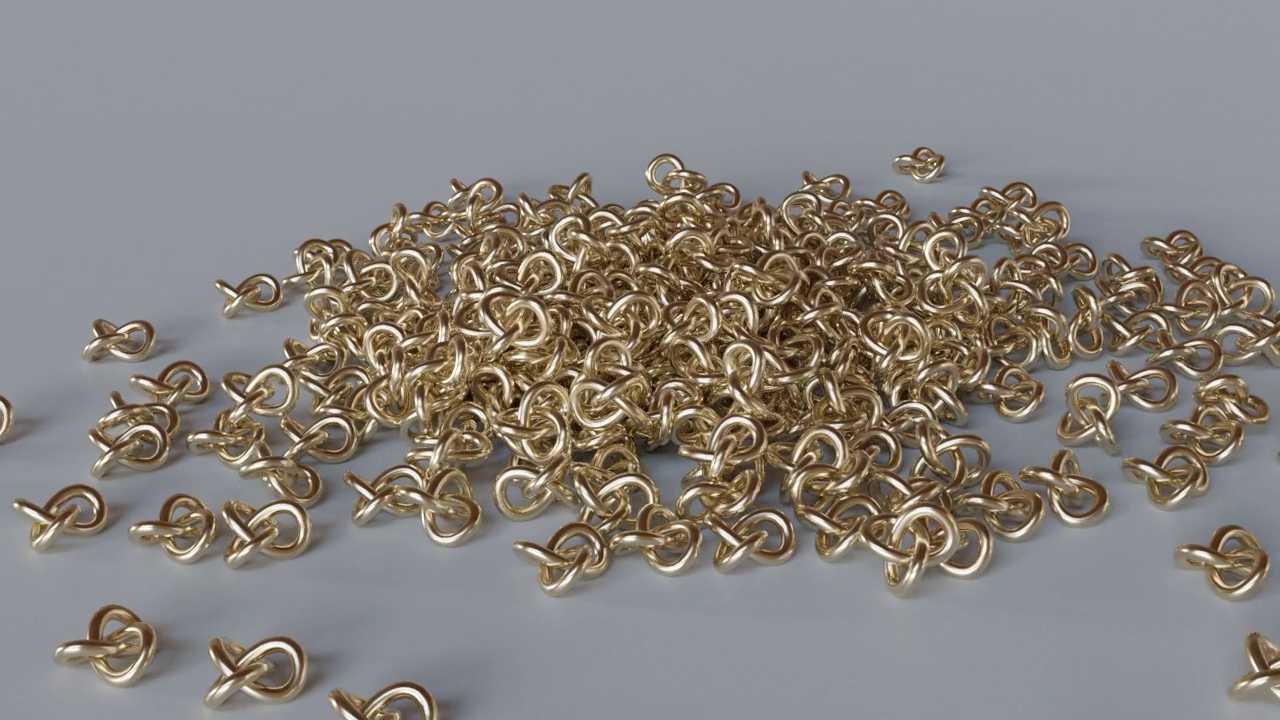}\\
\includegraphics[width=\rolloutwidth]{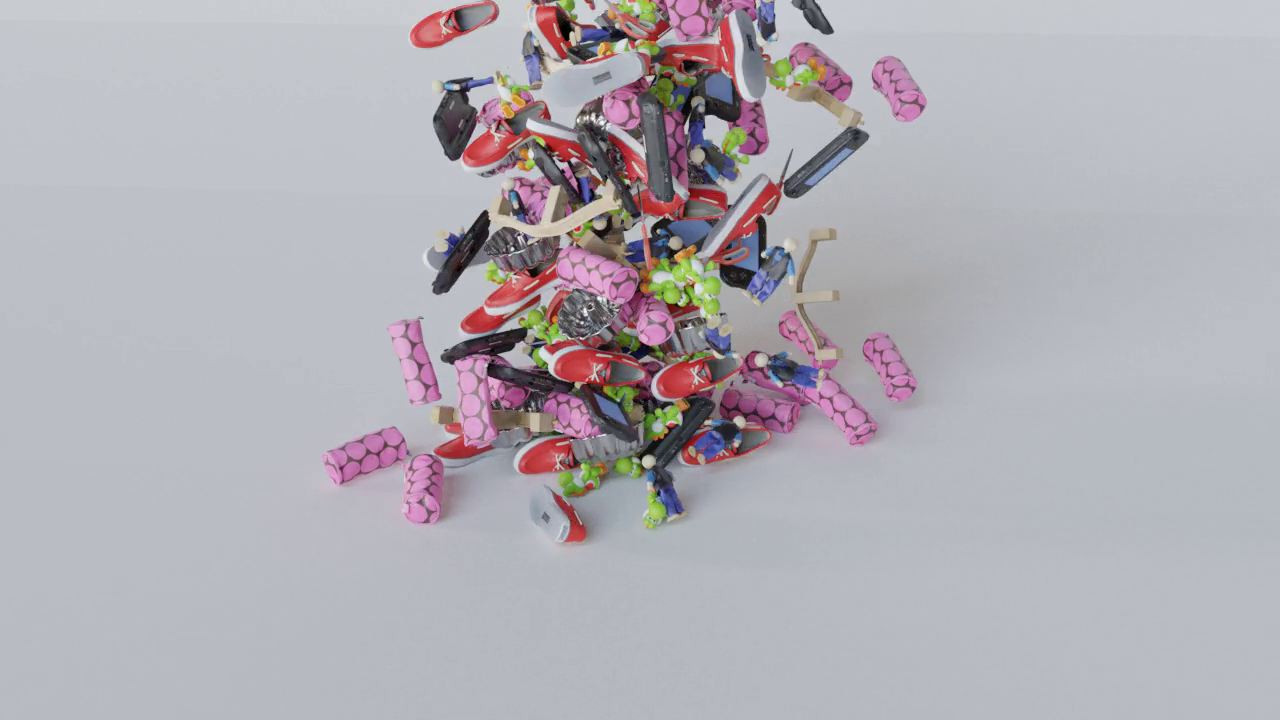}
\includegraphics[width=\rolloutwidth]{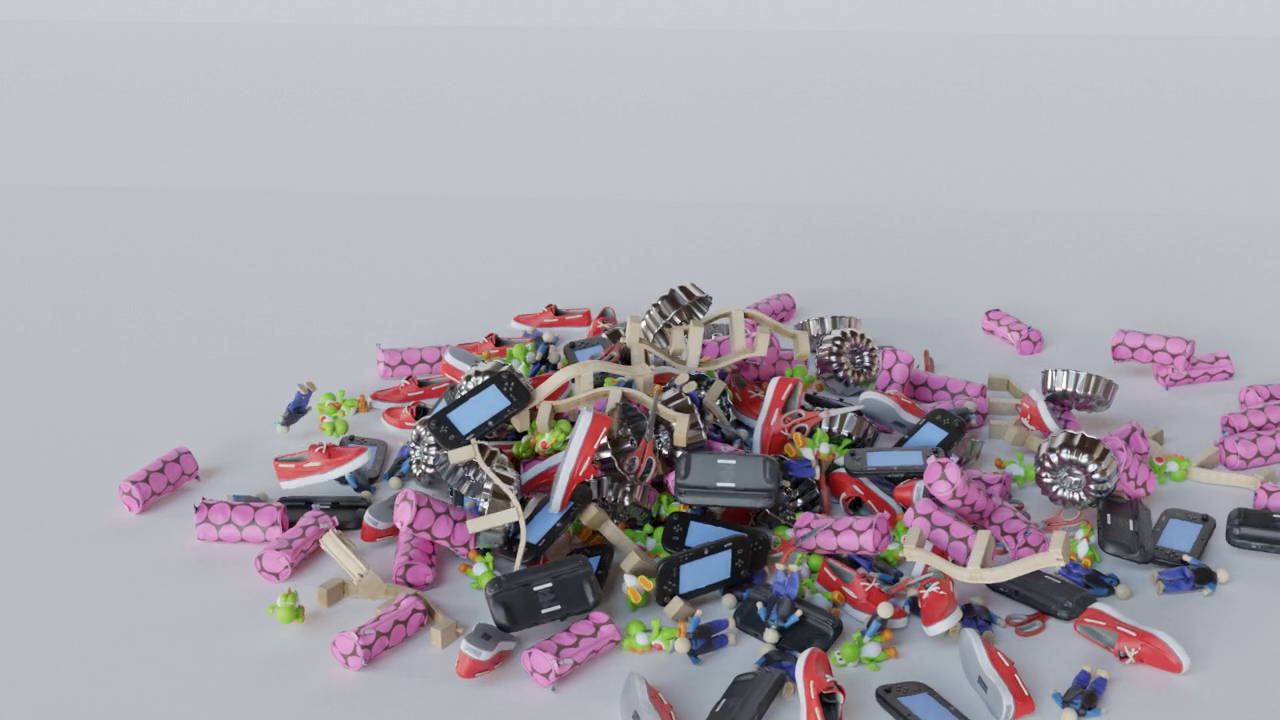}
\includegraphics[width=\rolloutwidth]{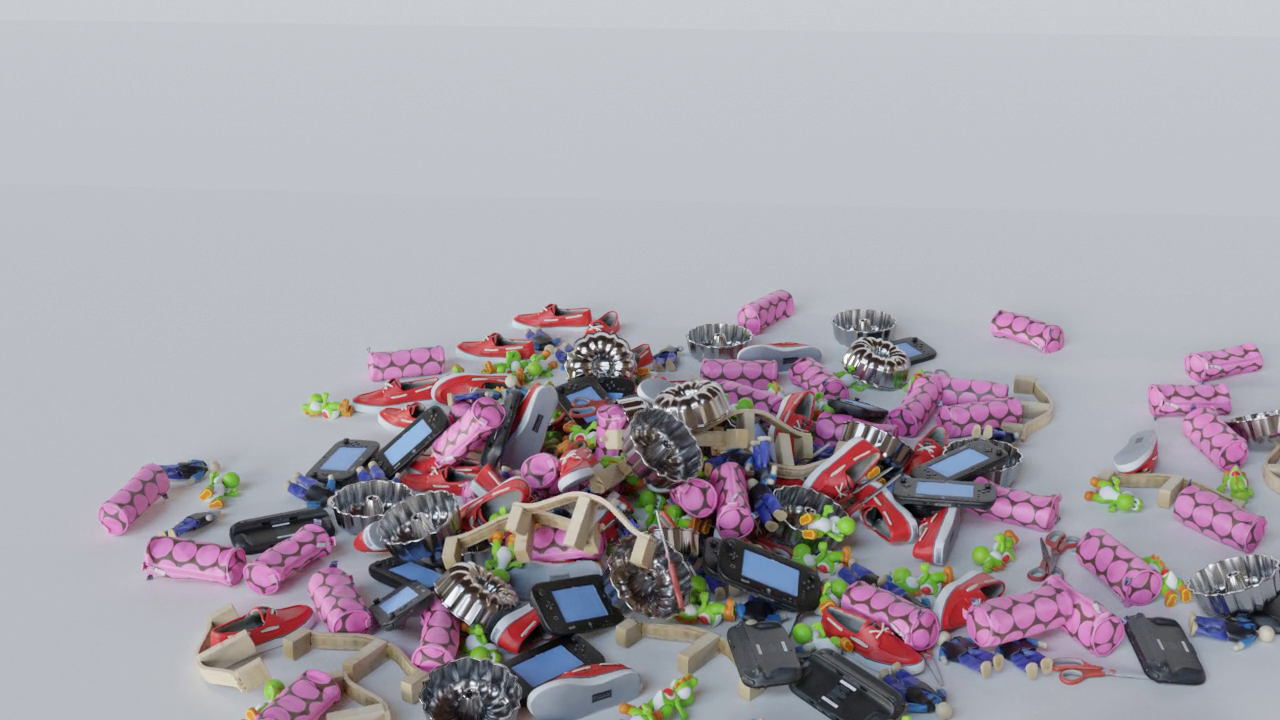}\
\caption{Examples of rollouts from SDF-Sim scaled to large simulations, all simulated for 200 steps. (Top) 
270 knots from Movi-B, 384k nodes (Bottom) 380 objects from Movi-C, 1.1M nodes.}
\label{app:large_falling_pile}
\end{figure*}

\section{Datasets and custom simulations}

\subsection{Kubric dataset and evaluations \cite{greff2021kubric}}
\label{app:mesh_sizes}

We use Movi-B and Movi-C datasets from Kubric \cite{greff2021kubric} for  small-scale evaluation and fair comparison to baselines \citep{allen2023fignet} and \citep{fignetplus}.
Simulations in Movi-B and Movi-C datasets contain between 3 and 10 objects per scene. In all simulations objects are thrown on the floor towards the center of the scene. The simulations consist of 100 time steps, or 2 seconds of the simulation time. See the visualisations of Movi-B shapes in Figure \ref{app:sdf_ablations_all_sdfs} and selected shapes from Movi-C in Figure \ref{app:movic_examples_learned}.

In both Movi-B and Movi-C we use 1500 trajectories for training, 100 for validation and 100 for test, as provided in the original Kubric dataset. Kubric dataset is distributed according to the Apache 2.0 license\footnote{\url{https://github.com/google-research/kubric/blob/main/LICENSE}}.

\begin{table}[h!] 
\centering
\begin{tabular}{lcccccc}
  & \# object types & \multicolumn{2}{c}{\# nodes per object} &   \multicolumn{2}{c}{\# triangles per object}\\
&  & mean & max & mean & max \\
\toprule
Movi-B  & 11 & 3,483 & 13,952 & 6,938 & 28,224 \\
Movi-C & 930 & 8,982 & 84,950 & 15,955 & 101,142 \\
\end{tabular}
\vspace{0.1cm}
\caption{Sizes of the \textbf{detailed meshes} used to train SDFs.}
\end{table}

\begin{table}[h!] 
\centering
\begin{tabular}{lcccccc}
  & \# object types & \multicolumn{2}{c}{\# nodes per object} &   \multicolumn{2}{c}{\# triangles per object}\\
& & mean & max & mean & max \\
\toprule
Movi-B  & 11 & 453 & 1,142 & 841 & 2,160 \\
Movi-C & 930 & 543 & 1,908 & 1,002 &  3,324  \\
\end{tabular}
\vspace{0.1cm}
\caption{Sizes of the \textbf{collision meshes} used for simulation.}
\end{table}











\begin{figure}[h!]
\centering
\begin{subfigure}[b]{0.69\textwidth}
\includegraphics[width=\textwidth]{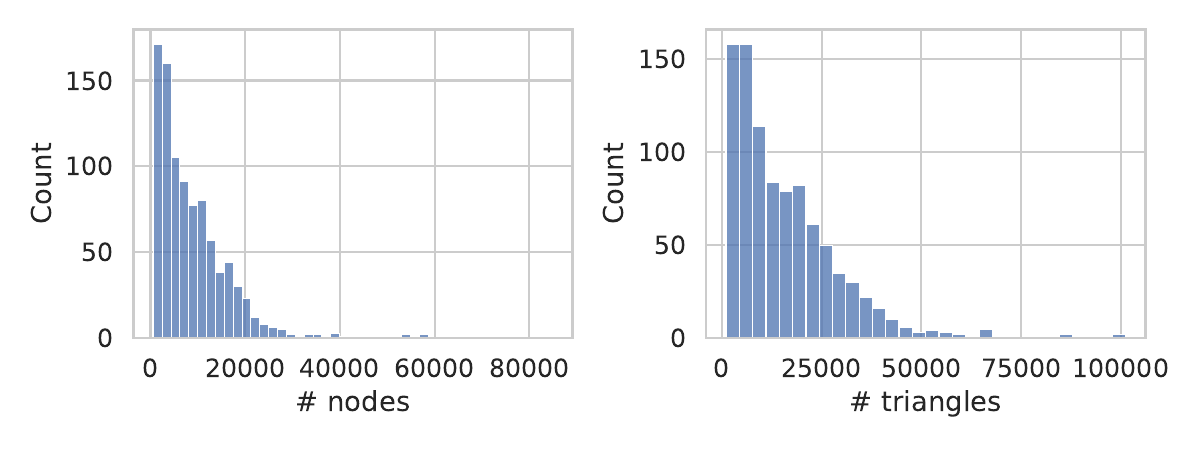}
\caption{Detailed meshes used for SDF training}
\end{subfigure}
\begin{subfigure}[b]{0.69\textwidth}
\includegraphics[width=\textwidth]{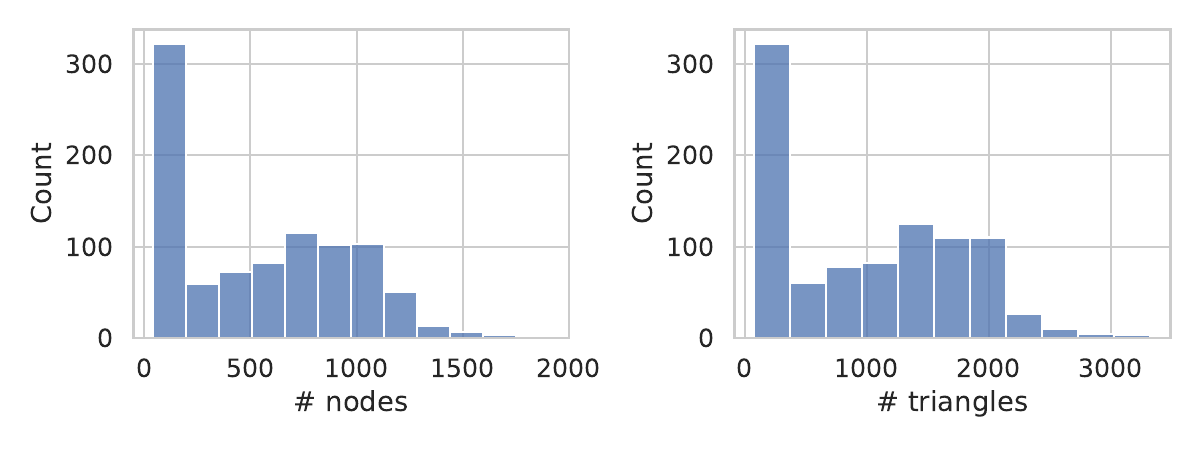}
\caption{Collision meshes used for simulation}
\end{subfigure}
\caption{Distribution of mesh sizes used in Movi-C (930 objects).}
\label{fig:mesh_sizes}
\end{figure}

\FloatBarrier

\subsection{Large scenes and vision}

For evaluations on large-scale scenes, we create several custom scenes in Blender with large number of objects. For each simulation, we arrange the objects in a stacked formation above the floor, to create a large number of collisions. Note that modelling a falling stack of objects is an extremely hard task even for classic simulators due to a large number of collisions.

\textbf{Spheres-in-Bowl}. For evaluations in section \ref{sec:scaling_spheres}, we create a scene with variable number of spheres (from 1 to 512) falling into a bowl. Spheres are objects from Movi-B, with 64 nodes each, resized to 0.52 meters in diameter. A bowl is an object from Movi-C. 

\textbf{Falling shoes} (Figure \ref{fig:large_falling_pile}). We use 300 identical shoe objects from Movi-C. This simulation contains 851k nodes and 200 frames

\textbf{Falling knots} (Figure \ref{app:large_falling_pile} top). We use 270 torus-knots from Movi-B. The resulting simulation has 384k nodes and 200 frames.

\textbf{Heaps of Stuff} (Figure \ref{app:large_falling_pile} bottom). We use 380 various objects from Movi-C . The scene consists of 1.1M nodes and is simulated for 400 time steps. 

\FloatBarrier

\section{SDF-Sim simulator architecture and training}
\label{app:training}

\paragraph{Architecture}

Following \cite{allen2023fignet}, we use a graph network with node and edge updates without global pooling. For node and edge updates, we use MLPs with 2 hidden layers and 128 units each with LayerNorm. GNN consists of 10 message-passing steps.

We add a small amount gaussian noise to the inputs in $\mathcal{N}(0, 0.003)$ to help the network to correct  inaccuracies from predictions in the previous steps and guarantee stable rollouts.

We normalise both inputs and outputs to zero-mean and unit-variance. The input graph is constructed using a history of positions for each node for the three latest time steps  $[\mathbf{n}_k^{t-2}, \mathbf{n}_k^{t-1}, \mathbf{n}_k^t]$ that we convert to a history of two previous velocities $[\mathbf{n}_k^{t-1} - \mathbf{n}_k^{t-2}, \mathbf{n}_k^t - \mathbf{n}_k^{t-1}]$. The model predicts the acceleration for each node $\hat{\mathbf{a}}_k^{t+1}$.

\paragraph{Loss}
We train the model on 1-step prediction task.  To construct a loss, we estimate ground-truth accelerations via finite difference: $\mathbf{a}_k^{t+1} = (\mathbf{n}_k^{t+1} - \mathbf{n}_k^{t}) - (\mathbf{n}_k^{t} - \mathbf{n}_k^{t-1}) = \mathbf{n}_k^{t+1} - 2  \mathbf{n}_k^{t} + \mathbf{n}_k^{t-1}$. We use mean-squared error loss on the accelerations per node (normalised as mentioned above) to train the model. Note that the loss is computed per-node rather than per object.

\paragraph{Generating a rollout}
To generate a rollout of multiple steps, we convert the predicted the per-node accelerations $\hat{\mathbf{a}}_k^{t+1}$\footnote{Here we use a "hat" $\hat{.}$ notation to denote the quantities that are estimated or predicted as opposed to the ground-truth values.} into the next simulation state (positions and translations of the objects). To do so, we estimate the positions of each node at the next time point as $\hat{\mathbf{n}_k}^{t+1} = \hat{\mathbf{a}}_k^{t+1}  + 2 \mathbf{n}_k^{t} - \mathbf{n}_k^{t-1}$. Finally, we use shape matching, similarly to \cite{allen2023fignet} to estimate the positions and rotation of the objects from the node positions.

\paragraph{Metrics}

Following \cite{allen2023fignet}, we report translation and rotation error between the predicted and ground-truth rollouts of 50 steps.
For translation, we report root mean-squared error over the object translation $\mathbf{o}_i^t$. For rotation, we report the mean angle between the predicted and ground-truth object rotations. Translation vector and rotation quaternions per object are computed from predicted node positions via shape matching, as described above. As in prior work \citep{allen2023fignet}, translation and rotation accuracy is reported on rollouts of 50 time steps, while memory and runtime metrics are reported on the full rollouts of 100 steps.

\paragraph{Training time and hardware} We train the network for 1M steps on 8 A100 GPUs devices with a batch size of 8. We use Adam optimizer, and an an exponential learning rate decay from 1e-3 to 1e-4. The approximate training time is ~5 days.

\section{Learning SDFs}
\label{sec_app:sdf_training}

\subsection{Training SDFs from meshes}
For our experiments, we pre-train the SDFs on each object that will be used in the simulation. For SDF training, we use the meshes in the canonical pose centered around the [0,0,0] point, rescaled so that the maximum vertex coordinate is 0.5. Thus, the entire object mesh is located within [-0.5, 0.5] interval for each axis.

\paragraph{Sampling query points} We sample query points $\textbf{y}$ on and near the surface of the mesh to use for SDF training.
First, we take the mesh nodes $\mathcal{V}_S$ and sample more points on the surface of the mesh uniformly using \texttt{sample\_points\_uniformly} function in open3D library. We sample 10,000 points per sq. meter of the object surface area. Then, we add gaussian noise with $\sigma=0.1$ to the surface points. We use both surface points and noisy points as query points for training. We re-sample the noisy points every 100 steps of training.

\paragraph{Computing ground-truth SDF}

We compute the ground-truth SDF $d$ to the object mesh for all query points. First,
for each query point $\mathbf{y}$, we use a classic Boundary Vector Hierarchy (BVH) \cite{bvh} method to find the closest triangle on the object mesh and then calculate the closest point $\mathbf{y}^*$ on the triangle. The ground-truth distance from the point $\mathbf{y}$ to the mesh is $||\mathbf{y} - \mathbf{y}^* ||$.
Finally, we use the normal $\mathbf{N}_f$ of the closest triangle (pointing outside of the object) to estimate the SDF sign. We compute the dot product $\mathbf{N}_f \cdot (\mathbf{y} - \mathbf{y}^*)$. If the dot product is positive, the point is outside the object and the SDF should have a positive sign, and if it is negative, it is inside the object and its SDF should have a negative sign.
The ground-truth SDF $d$ is therefore $||\mathbf{y} - \mathbf{y}^* ||$ multiplied by the sign.

\paragraph{Training SDF} We train the SDF function $f_{\theta}(\mathbf{y})$ by supervising it using the ground-truth SDF $d$.
We use the L2 loss on the signed distances $(d - f_{\theta}(\mathbf{y}))^2$, as well as the L2 loss on the absolute distances  $( |d| - |f_{\theta}(\mathbf{y})| )^2$.
The reason is that we found that sign estimation can be unstable if vector $(\mathbf{y} - \mathbf{y}^*)$ is close to being orthogonal to the triangle normal vector $\mathbf{n}_f$.  Therefore, we use the loss on \textit{unsigned} distances on all query points, but the loss on the \textit{signed} distances is only used on the points where the sign can be reliably estimated. Specifically, we use \textit{signed}-distance loss on the points where the angle between $(\mathbf{y} - \mathbf{y}^*)$ and $\mathbf{n}_f$ is less than 75 degrees or more than 105 degrees.

\paragraph{SDF architecture}
Unless otherwise stated, the learned SDFs consisted of MLPs with 8 layers and 128 hidden units. The models were trained for 400K steps and a learning rate of $10^-5$ for Kubric Meshes. For the Garden Vase scene, MLPs with 8 layers and 256 units were used, but training details were otherwise the same. For quantitative results in Figure \ref{fig:accuracy_comparison}, we run the all the model with 3 seeds and report the mean and the 95\% confidence interval.

\paragraph{SDF training time and hardware}

For all meshes in Movi-C, we trained the MLPs for 400k iterations, which requires 18 hours of training time on average on a single V100 GPU. It is likely that this training time could be reduced, as we did not aim to minimize the computational cost of the SDFs in this work. Instead, we chose to train the SDF to maximize the quality for complex Movi-C meshes that have 8,982 nodes and 15,955 triangles on average. 


\paragraph{Other approaches tested}
In NeRF literature a common regularisation to train SDFs is the Eikonal regularisation that enforces unit-norm of the SDF gradient. However, we found that Eikonal regularisation leads to over-smoothed sphere-like shapes and did not use it in our training.
Furthermore, we find that our approach of directly fitting the distances already results in unit-norm-gradient and does not require an additional regularisation term.  We also tested another common technique of periodic encodings as the MLP inputs, however, we found that it leads to checkerboard-like artifacts for points outside of the object and did not use this method.

\subsection{Training SDFs from vision}
\label{app:sdf_vision}

For an experiment in section \ref{sec:sdf_vision}, we take a Garden Scene from \cite{mipnerf360}, represented as a set of 2D images with a 360-degree view of the scene. We train a VolSDF model \cite{volsdf} to distill an SDF from this scene. VolSDF leverages a volumetric representation that explicitly encodes the volume density as a transformed SDF. This explicit formulation provides an inductive bias over the geometry of an object and a built-in preference for capturing smooth and clean shapes. 

VolSDF already represents a learned SDF and, in principle, can be directly used in SDF-Sim. However, to simplify the pipeline and have a consistent SDF architecture for all objects, we train another learned SDF, as described in Section \ref{sec_app:sdf_training}. To do so, we extract a high quality mesh (with over 80k vertices) of the table and the vase from VolSDF using the Marching Cubes algorithm \citep{marching_cubes}. Then we train a learned SDF with the architecture described in section \ref{sec_app:sdf_training} and use it in the SDF-Sim.



\clearpage

\section{Additional results}

\subsection{Quantitative Results on small benchmark datasets.}
\label{app:quantitative_results}

\setcounter{figure}{0}
\renewcommand{\thefigure}{S\arabic{figure}}


\begin{table}[hb!] 
\centering
\textbf{Movi-B}\\
\small
\vspace{0.3cm}
\begin{tabular}{lcc}
\textbf{Model} & \textbf{Translation RMSE (m)} & \textbf{Rotation Err (deg)}\\
\toprule
DPI & 0.368 $\pm$ 0.057 & 26.928 $\pm$ 2.740 \\
MGN-LargeRadius & 0.460 $\pm$ 0.045&  26.342 $\pm$ 1.397 \\
MGN &  0.538 $\pm$ 0.035 & 26.914 $\pm$ 0.783 \\
FIGNet & 0.14 $\pm$ 0.0087 & \textbf{15.0} $\pm$ 0.675  \\
FIGNet* & \textbf{0.13} $\pm$ 0.008 & 15.94 $\pm$ 0.8849  \\
\hline
SDF-Sim (ours) & 0.16 $\pm$ 0.0124 & 18.03 $\pm$ 0.9433 \\
\vspace{0.4cm}
\end{tabular}
\begin{tabular}{lcccc}
\textbf{Model} & \textbf{\# Collision Edges} & \textbf{\# Graph Edges} & \textbf{Peak Memory}  & \textbf{Runtime per step}\\
& & & \textbf{(MiB)}  & \textbf{(ms)}\\
\toprule
DPI & 2250.688 $\pm$ 65 & - & - & 0.145 $\pm$ 0.009 \\
MGN-LargeRadius & 1797.985 $\pm$ 60 & - & - &  0.218 $\pm$ 0.033 \\
MGN & 34.367 $\pm$ 4 & - & - & 0.175 $\pm$ 0.018 \\
FIGNet & 1360.99 $\pm$ 196 & 24569.59 $\pm$ 938 & 94.45 $\pm$ 2.9698 & 0.23 $\pm$ 0.0044 \\
FIGNet* & 1409.87 $\pm$ 267 & 8613.05 $\pm$ 512 & 63.83 $\pm$ 3.7489 & 0.22 $\pm$ 0.0077 \\
\hline
SDF-Sim (ours) & \textbf{201.14} $\pm$ 10 & \textbf{6195.58} $\pm$ 193 & \textbf{38.8} $\pm$ 0.5711 & \textbf{0.14} $\pm$ 0.004 \\
\vspace{0.1cm}
\end{tabular}
\caption{Quantitative comparison of SDF-Sim and the baselines on a \textbf{Movi-B} dataset. This table presents the numbers for the Figure \ref{fig:accuracy_comparison} in the main text. The error bars show the 95\% confidence interval from running all models with 3 random seeds. DPI, MGN-LargeRadius and MGN results are reported by \cite{allen2023fignet}. Results marked as '-' were not reported by the previous work \cite{allen2023fignet}}.
\label{app:accuracy_comparison_movi_b}
\end{table}


\begin{table}[hb!] 
\centering
\textbf{Movi-C}\\
\small
\vspace{0.3cm}
\begin{tabular}{lcc}
\textbf{Model} & \textbf{Translation RMSE (m)} & \textbf{Rotation Err (deg)}\\
\toprule
DPI-Reimplemented* & \multicolumn{2}{c}{OOM}\\
MGN-LargeRadius* & \multicolumn{2}{c}{OOM}\\
MGN* & \multicolumn{2}{c}{OOM} \\
FIGNet & \multicolumn{2}{c}{OOM} \\
FIGNet* & \textbf{0.18} $\pm$ 0.0097 & \textbf{19.68} $\pm$ 0.6483 \\
\hline
SDF-Sim (ours) & 0.24 $\pm$ 0.0151 & 23.15 $\pm$ 0.668 \\
\vspace{0.4cm}
\end{tabular}
\begin{tabular}{lcccc}
\textbf{Model} & \textbf{\# Collision Edges} & \textbf{\# Graph Edges} & \textbf{Peak Memory}  & \textbf{Runtime per step}\\
& & & \textbf{(MiB)}  & \textbf{(ms)}\\
\toprule
DPI-Reimplemented* & \multicolumn{4}{c}{OOM} \\
MGN-LargeRadius* & \multicolumn{4}{c}{OOM}  \\
MGN*  &  \multicolumn{4}{c}{OOM}  \\
FIGNet  & \multicolumn{4}{c}{OOM}  \\
FIGNet* & 2220.96 $\pm$ 469 & 11502.18 $\pm$ 944 & 75.87 $\pm$ 9.587 & 0.30 $\pm$ 0.0163 \\
\hline
SDF-Sim (ours) & \textbf{300.46} $\pm$ 34 & \textbf{7661.17} $\pm$ 266 & \textbf{43.35} $\pm$ 0.7466 & \textbf{0.17} $\pm$ 0.0038 \\
\vspace{0.1cm}
\end{tabular}
\caption{Quantitative comparison of SDF-Sim and the baselines on a \textbf{Movi-C} dataset. This table presents the numbers for the Figure \ref{fig:accuracy_comparison} in the main text. The error bars show the 95\% confidence interval from running all models with 3 random seeds. DPI, MGN-LargeRadius and MGN and FIGNet baselines run OOM on the Movi-C simulations and are not reported.}.
\label{app:accuracy_comparison_movi_c}
\end{table}

\clearpage

\subsection{Additional metrics on SDF quality}
\label{app:sdf_quality}

\begin{figure}[hb!]
\centering
\includegraphics[trim={0 0 0 10},clip,width=0.7\textwidth]{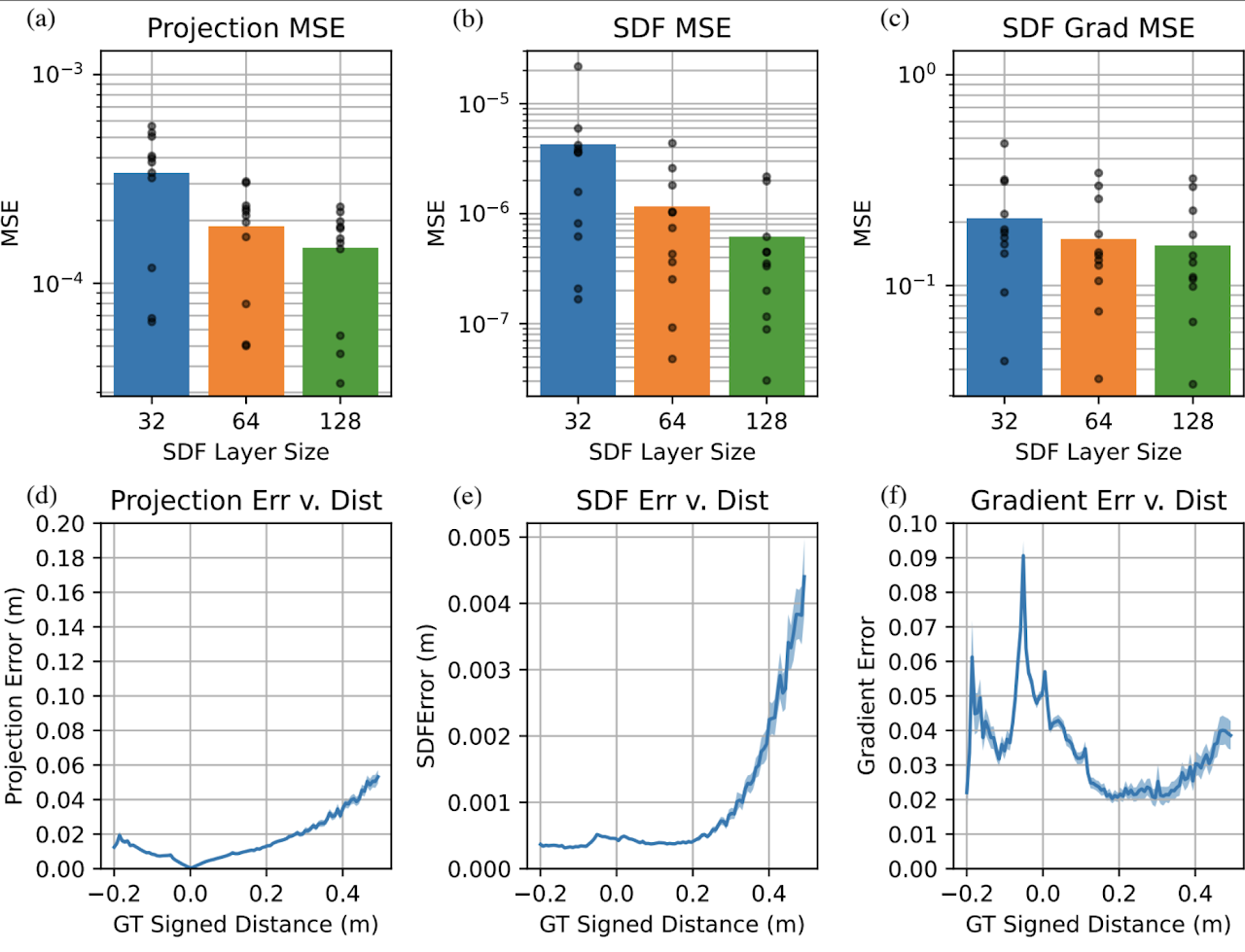}
\caption{\textbf{Additional metrics on SDF quality}\\
\textbf{(a-c)} Metrics computed across learned SDF models of different sizes for different objects in Movi-B. Each black marker denotes a different Movi-B shape. For each metric, loss decreases as we increase the size of the model. \\
\textbf{(a)} MSE between the projected surface points predicted from the learned SDFs and the true closest point on the surface.\\
\textbf{(b)} MSE between the signed distance outputted by the learned SDF and the true signed distance.\\
\textbf{(c)} MSE between the gradient of the learned SDF and the gradient of the ground truth SDF function.\\
\\ \textbf{(d-f) SDF evaluation metrics as a function of distance from the object surface}. SDF error,  projection error and the gradient errors remain constant up to distance ~0.2 from the object surface and start to increase afterwards. Generally all errors remain low in comparison to the object size of 0.5 meter. \\
\textbf{(d)} Error as distance between the learned SDF projected surface point and the true closest surface point, plotted as a function of signed distance from the surface.\\
\textbf{(e)} Error as the absolute difference in magnitude between the learned SDF prediction and ground truth SDF, plotted as a function of signed distance from the surface.\\
\textbf{(f)} Error as the cosine distance between the gradient of learned SDF and the ground truth SDF gradient.}
\label{fig:sdf_quality_suppl}
\end{figure}

\FloatBarrier



\FloatBarrier
\clearpage

\subsection{Runtime versus number of nodes in simulation graph.}

We measure the time to execute one step of simulation as a function of the input graph on different scenes in Movi-C dataset. For both  SDF-Sim and FIGNet*, the runtime grows linearly with the number of the nodes in the graph. The runtime of SDF-Sim is consistently lower than FIGNet*. On large meshes with $\sim$8000 nodes and $\sim$18000  peak edges, SDF-Sim has a 52\% smaller runtime compared to the FIGNet* baseline.
 
\begin{figure}[h!]
\centering
\includegraphics[width=0.29\textwidth]{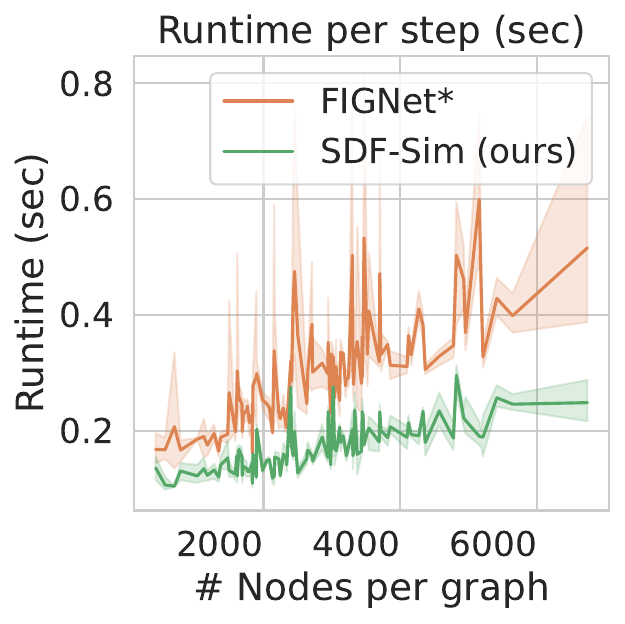}
\caption{\textbf{Runtime per simulation step} w.r.t. the total number of nodes in the scene. The runtime evaluations were performed on the Movi-C test set on Nvidia A100 GPU. FIGNet baseline runs OOM on Movi-C and is not shown here.}
\label{fig:runtime_comparison}
\end{figure}

\FloatBarrier

\subsection{Memory of the SDF versus a mesh}
\label{sec:sdf_mesh_memory}

\begin{figure}[h!]
\centering
\includegraphics[width=0.5\textwidth]{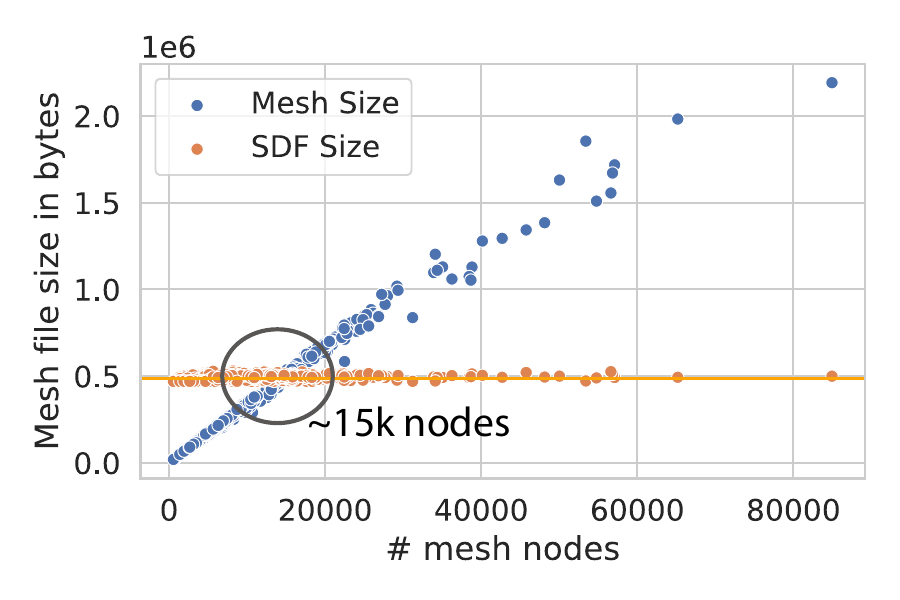}
\caption{
\textbf{Memory footprint of storing the mesh versus storing the SDF weights} w.r.t. the number of mesh nodes, evaluated on 930 objects from Movi-C.}
\label{fig:sdf_mesh_memory}
\end{figure}

A key advantage of learned SDFs is that they provide a compact representation of complex object shapes compared to meshes.
In Figure \ref{fig:sdf_mesh_memory}, we compare the sizes of meshes from Movi-C dataset (red) to the sizes of the SDF model parameters (purple) in MiB, assuming both are stored as pickle files on disk. While the size of the meshes grows with the number of mesh nodes, memory requirements of the SDFs do not depend on the complexity of the object, as both simple and intricate shapes are modeled with the same size MLP. We see that our MLP with 8 layers of 128 units has similar memory demands to meshes with ~15,000 nodes, which is 4x less memory than the largest meshes in Movi-C.
Thus, SDF-Sim can potentially use less memory to store the object shapes: while FIGNet and FIGNet* require to store the entire object mesh, SDF-Sim needs only a compact SDF and a set surface nodes, which, as we will show in section \ref{sec:adaptive_sampling}, can be relatively small.

\FloatBarrier
\clearpage

\subsection{Re-sampling object surface nodes using an SDF}
\label{sec:adaptive_sampling}

\begin{figure}[h!]
     \begin{subfigure}[b]{0.13\textwidth}
         \captionsetup{justification=centering,singlelinecheck=false}
         \centering
         \caption{\footnotesize Detailed Mesh\\ K=21285}
         \includegraphics[trim={8cm 3cm 9cm 4cm},clip,width=\textwidth]{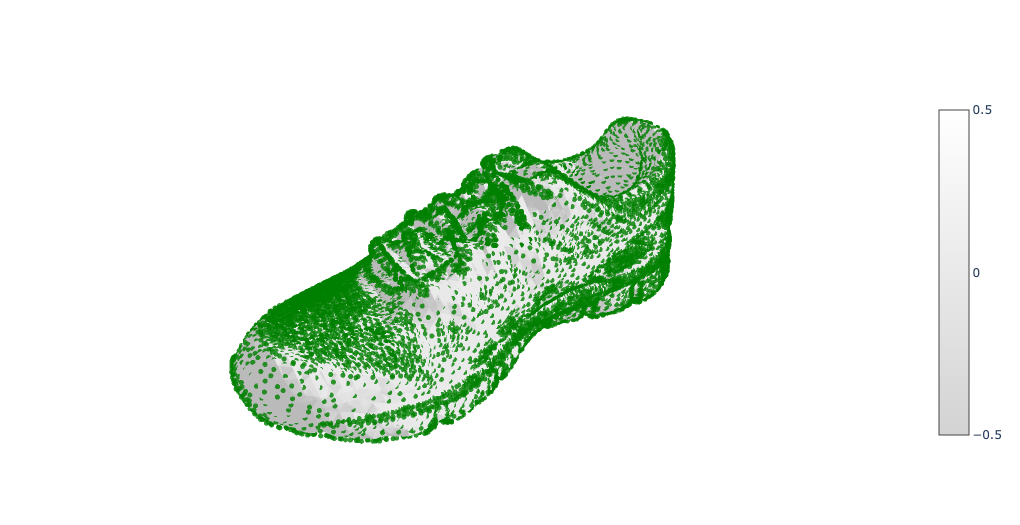}
         
    \end{subfigure}
     \begin{subfigure}[b]{0.13\textwidth}
        \captionsetup{justification=centering,singlelinecheck=false}
        \caption{\footnotesize Simulation Mesh \\K=759}
         \includegraphics[trim={7cm 2cm 9cm 4cm},clip,width=\textwidth]{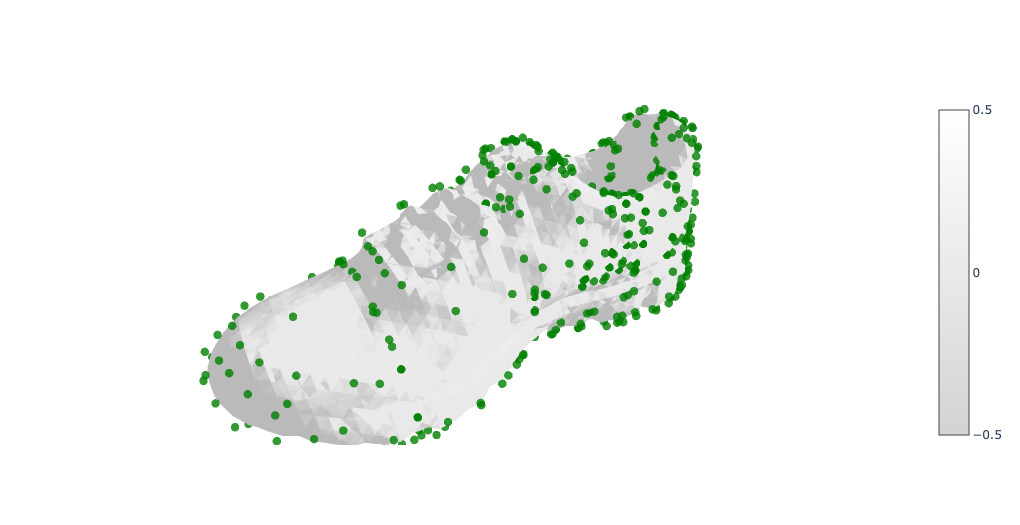}
    \end{subfigure}
     \begin{subfigure}[b]{0.13\textwidth}
        \captionsetup{justification=centering,singlelinecheck=false}
        \caption{\footnotesize Sampled from SDF K=676}
         \includegraphics[trim={7cm 2cm 9cm 4cm},clip,width=\textwidth]{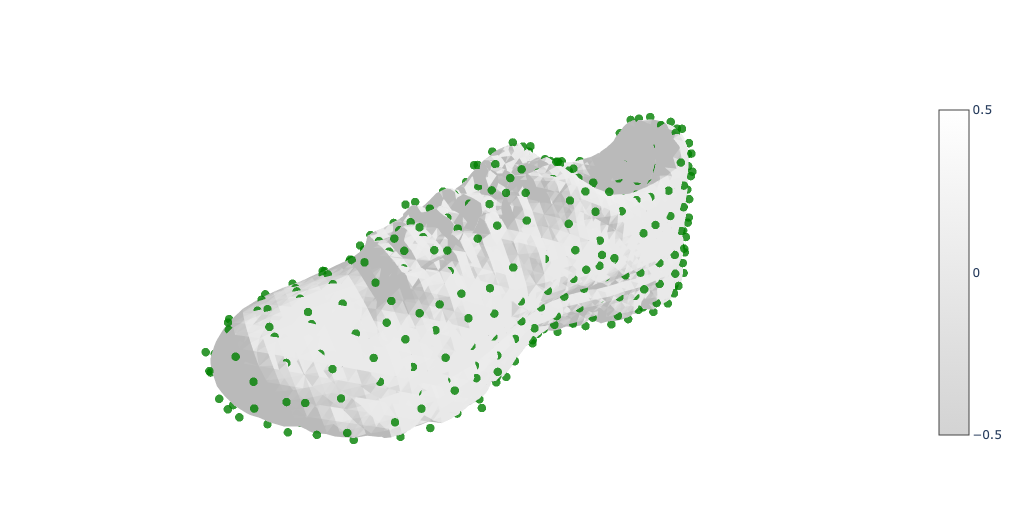}
     \end{subfigure}
     \centering
    \begin{subfigure}[b]{0.54\textwidth}
         \centering
         \includegraphics[width=\textwidth]{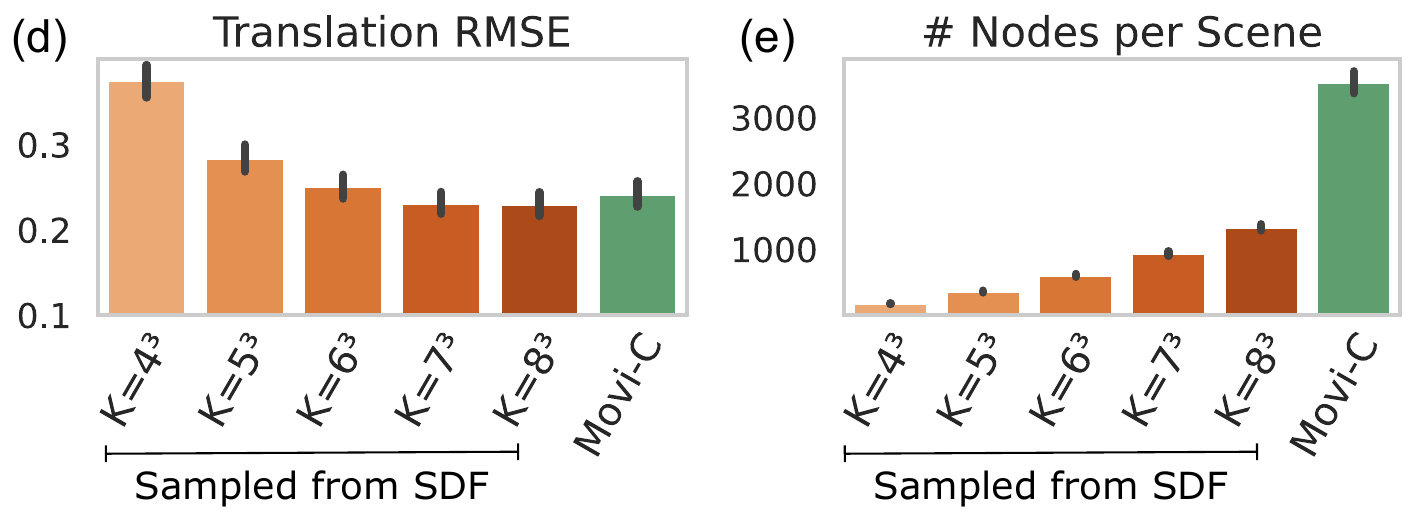}
     \end{subfigure}
    
         
    \caption{Shoe object from Movi-C, with vertices shown in green: (a) the original high-resolution mesh for rendering, (b) collision mesh for simulation, (c) nodes sampled on a grid and projected to the learned SDF).
    We can achieve favorable tradeoffs between error (d) and node count (e) by varying the dimensions $K$ of the sample grid (orange bars). The green bar represents SDF-Sim that uses the nodes from the original Movi-C collision mesh, as used elsewhere in the paper.}
    \label{fig:sdf_sampled_points}
\vspace{-0.3cm}
\end{figure}

Simulation accuracy heavily depends on the quality of the object shapes, typically a mesh. However, not all meshes are suitable for simulation. Object meshes, either created by artists or obtained by scanning a real-world object, often consist of thousands of nodes, and the nodes are concentrated around on the fine details of the shape (an example shown in Figure \ref{fig:sdf_sampled_points}(a)). 
To be able to simulate the objects, the shapes are often approximated by hand-crafted convex \emph{collision meshes} (Figure \ref{fig:sdf_sampled_points}(b)). But such meshes might not always exist, and might not be optimal for GNN-based simulators like FIGNet. However, mesh-based simulators like FIGNet, typically use these collision meshes, as changing a mesh while retaining a connected surface is not easy.

In contrast, SDF-Sim does not require a connected mesh; it relies only on surface nodes and the learned SDF to perform a simulation step. This gives us full flexibility to choose the number and the location of graph nodes.  Varying the number of surface nodes gives us a tighter handle controlling the tradeoff between accuracy, simulation speed and memory.

To create a new set of surface nodes, we can
\emph{sample} the surface nodes from a learned SDF. We initialize $K = N^3$ equally-spaced points in the volume of a cube which encompasses the object. We project these points to the object surface using Eq.~\ref{eq:sdf_proj}, with minor filtering (more details on this sampling procedure described below).
With this strategy, we get points that evenly cover the entire object surface (Figure \ref{fig:sdf_sampled_points}(c)). Finally, we take the SDF-Sim simulator trained on Movi-C and replace each object's mesh nodes with the ones sampled from an SDF \textit{at test time}.
Figure \ref{fig:accuracy_comparison} shows that with this sampling strategy, we can maintain the simulation accuracy using 73.5\% fewer nodes (hence smaller memory) compared to SDF-Sim with Movi-C collision meshes, used elsewhere in the paper. Specifically, in Figure \ref{fig:sdf_sampled_points}(d) the models with $K=7^3$ and $K=8^3$ (dark brown bars) have similar translation error (0.23) compared to the model with original node distribution (green bar, 0.24). However, in Figure \ref{fig:sdf_sampled_points}(e) the model $K=7^3$ requires only 934.19 nodes per scene on average (fourth orange column), while the model with the original Movi-C nodes needs 3530.13 nodes (green column) to perform the same simulation.

Here we perform this experiment at test time only, but it would be straightforward to also \textit{train} the SDF-Sim with the nodes sampled from an SDF, and experiment with sampling strategies for optimal prediction performance. This creates the opportunity to use a learned SDF as the sole representation of shape without ever requiring a mesh.

\paragraph{Details on sampling the surface nodes from an SDF}
We start with a uniform 3D grid of NxNxN points, located within an interval $\left[-0.7, 0.7\right]$, to ensure that the grid is larger than the object (recall that the object meshes were rescaled to $\left[-0.5, 0.5\right]$ before training an SDF). Placing the query points on a grid ensures that the corresponding projected points would cover the entire object surface, including concave parts like inner side of a shoe or a knot. Then, we project these points on the surface: $\mathbf{y^*} = \mathbf{y} - f_{\theta}(\mathbf{y}) \nabla f_{\theta}(\mathbf{y})$.
We generally find that the trained SDF functions are fairly accurate and the projected points are close to the surface. Nevertheless, we additionally filter the points $f_{\theta}(\mathbf{y^*}) < 0.01$ to ensure that $\mathbf{y^*}$ are exactly on the surface as it otherwise would impact the accuracy of the simulation. Finally, some areas of the shape might have clusters of points. We additionally downsample projected points using {\ttfamily voxel\_down\_sample} function from open3D library \cite{Open3D}, such that at most one point is present per voxel of size 0.025 to further reduce the number of points.

\FloatBarrier

\subsection{Quantifying the source of SDF-Sim error}

A natural question to ask is, where does the increased error (in comparison to FIGNet*) of SDF-Sim on Movi B/C comes from. A possible way to quantify this would be to train an SDF-Sim architecture using the accurate distances directly computed from a mesh. We note that correctly estimating \textit{signed} distances is actually hard and is subject to edge cases.
We ran an experiment where we train SDF-Sim on Movi-C with estimates of the \emph{unsigned} distance (UDF) computed from the mesh via BVH. We achieve similar Translation RMSE on Movi-C (0.23 with mesh-based UDFs versus 0.24 with learned SDF). While it is not a one-to-one comparison, this result does indicate that our method is relatively robust to SDF accuracy, and the learned simulator can learn to correct for minor inaccuracies.

\FloatBarrier

\clearpage

\section{Learned SDF visualisations}

\begin{figure*}[hb!]
\centering
\begin{subfigure}[b]{0.45\textwidth}
\centering
\caption{Learned SDF reconstructions}
\includegraphics[width=\textwidth]{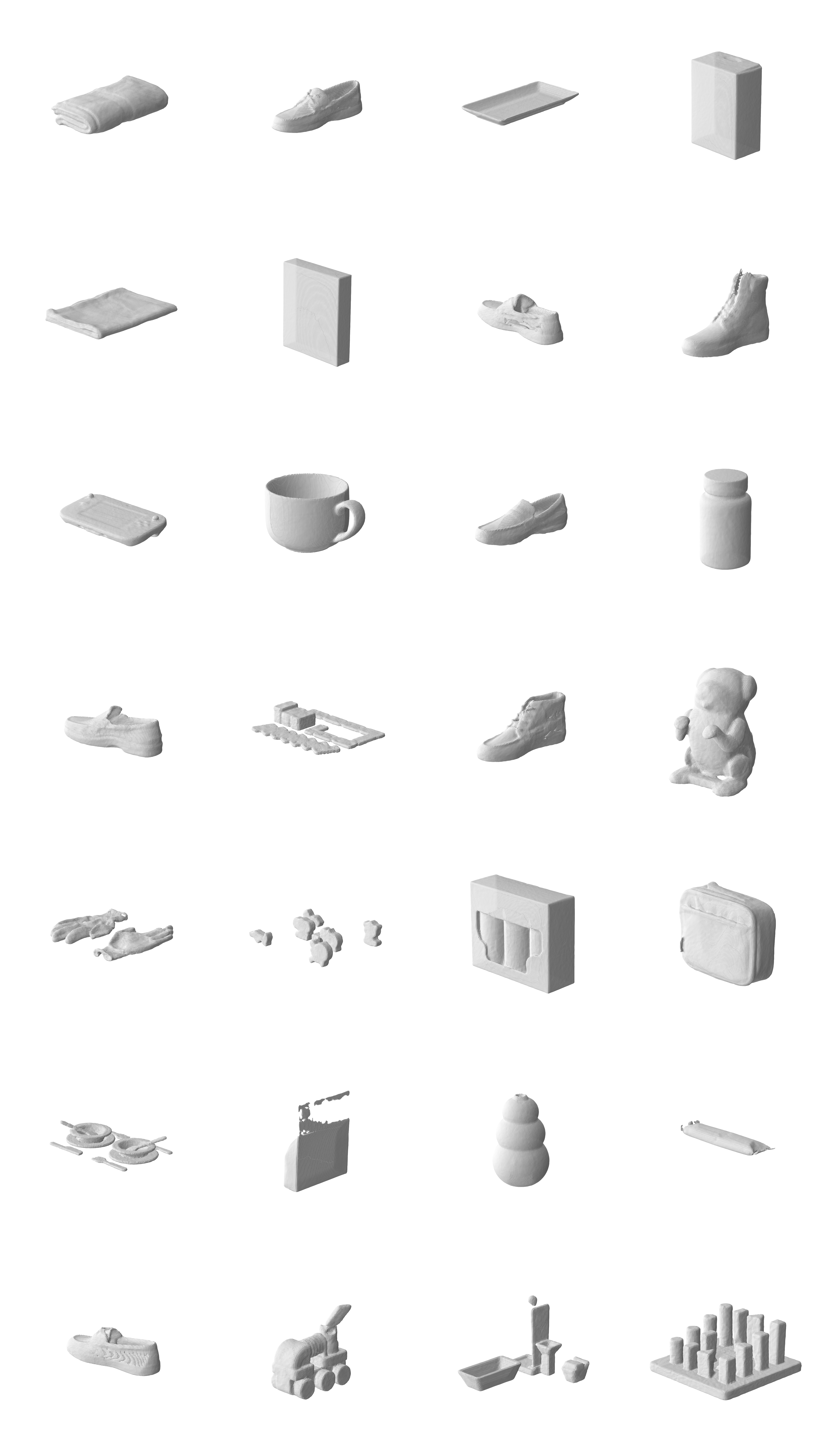}
\end{subfigure}
\unskip\ \vrule\ 
\begin{subfigure}[b]{0.45\textwidth}
\centering
\caption{Ground truth meshes}
\includegraphics[width=\textwidth]{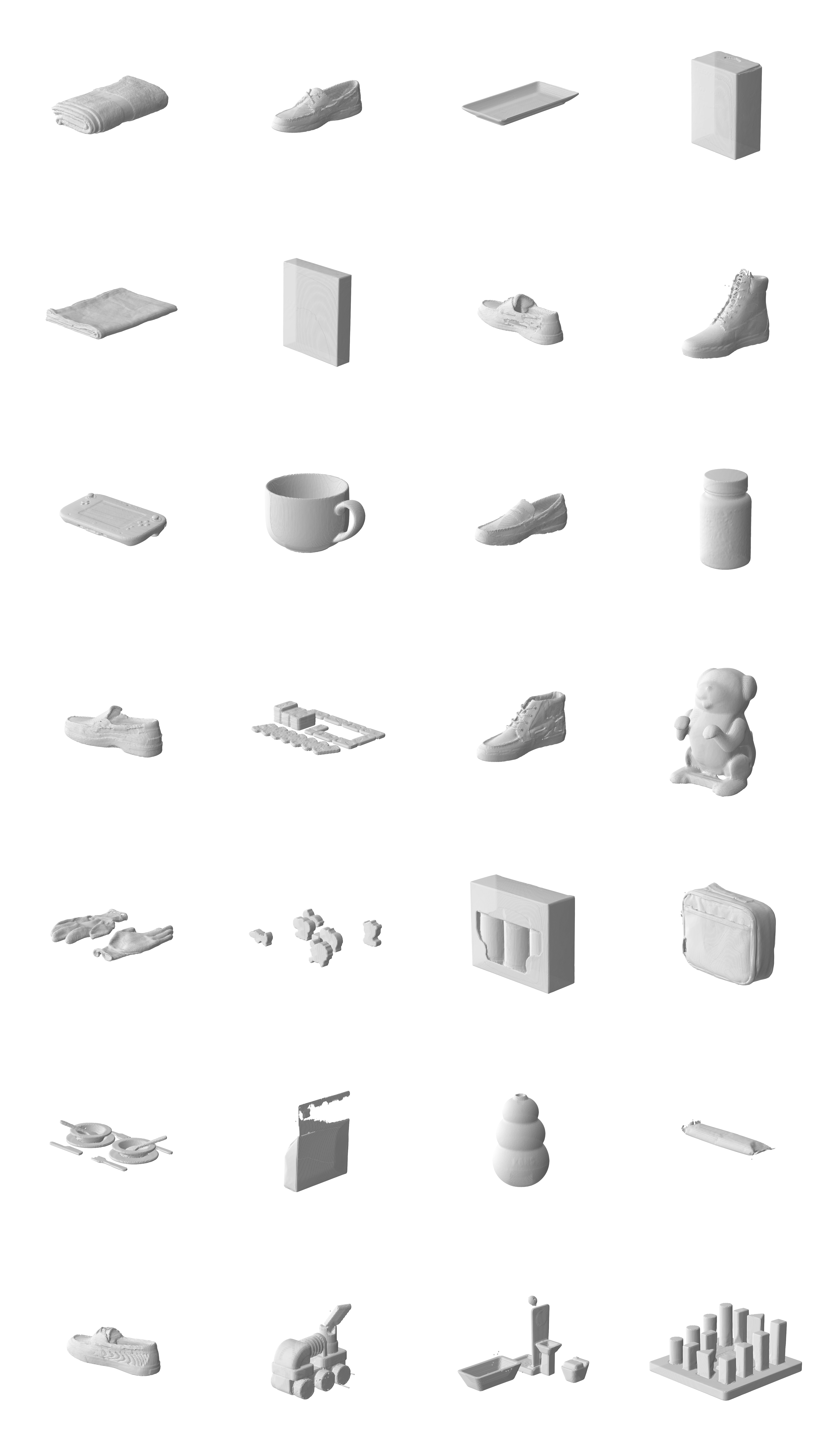}
\end{subfigure}
\caption{(a) Learned SDF reconstructions for randomly chosen Movi-C shapes. For visualisation purposes, we convert the object surface defined by a learned SDF: $\{\mathbf{y}: f_\theta(\mathbf{y})=0\}$ into a mesh using Marching Cubes and show this mesh here. (b) Corresponding ground-truth meshes.}
\label{app:movic_examples_learned}
\end{figure*}


\clearpage

\subsection{SDF shapes through training}

\begin{figure*}[hb!]
\centering
\includegraphics[width=0.6\textwidth]{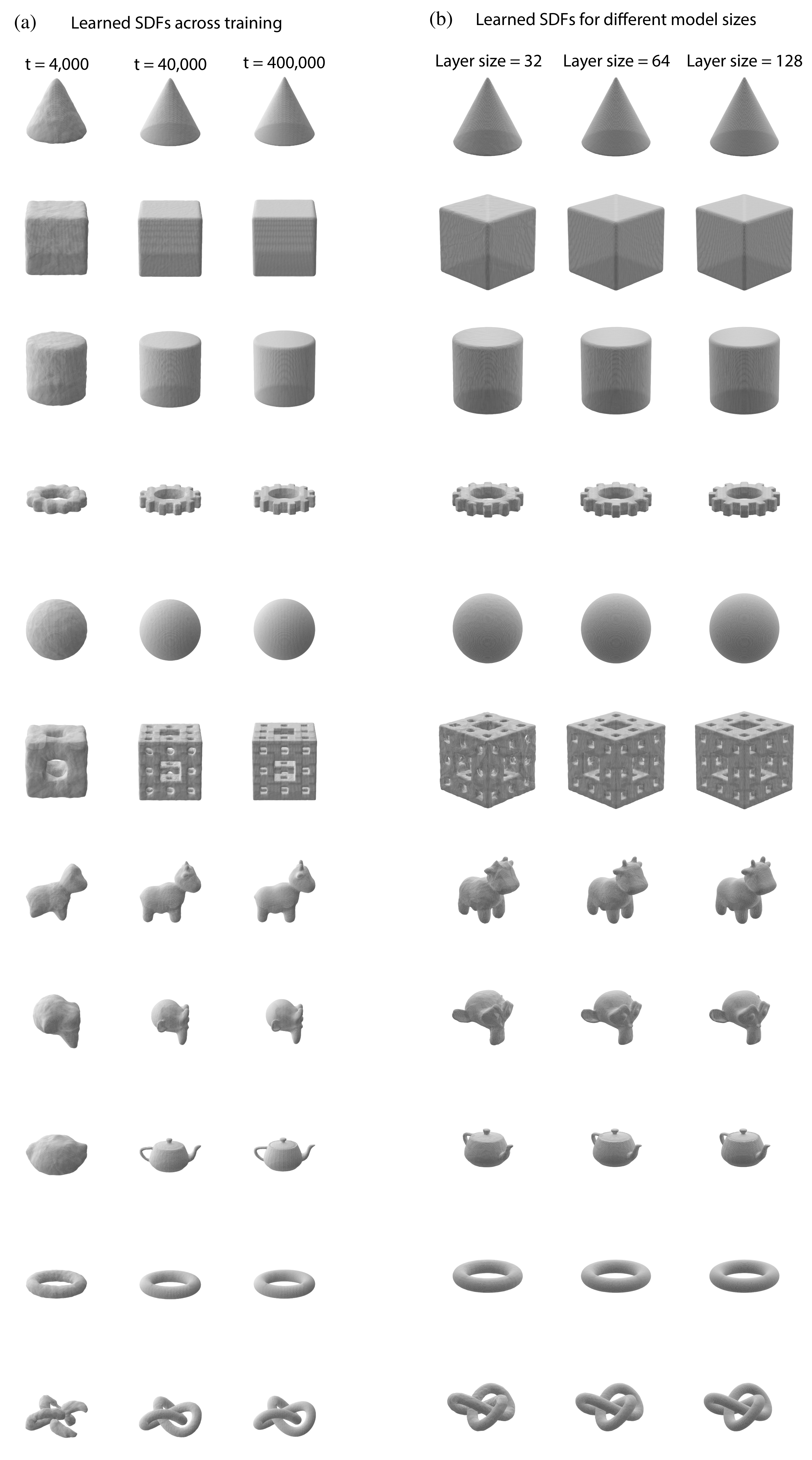}
\caption{(a) Learned SDFs for Movi-B visualized at different points during training. (b) Learned SDFs for Movi-B visualized for different MLP sizes (different number of units per layer, all MLPs have 8 layers). For visualisation purposes in both figures, we convert the object surface defined by a learned SDF: $\{\mathbf{y}: f_\theta(\mathbf{y})=0\}$ into a mesh using Marching Cubes.}
\label{app:sdf_ablations_all_sdfs}
\end{figure*}

As the SDF network has to encode the distances only for a single object, we find that a simple MLP is sufficient to encode complex shapes to the necessary extent for the simulation. Figure  \ref{app:sdf_ablations_all_sdfs}(a) demonstrates that the most accurate shapes are at 400k iterations of training, however already after 4000 training steps the network is able to capture the outlines of most shapes.
For simple shapes like cones or cubes, a model learns an accurate shape already by 8000 training iterations. Shapes with holes (like torus) can also be learned within the first 8000 iterations.
%
Although in this paper we chose to train the SDFs for 400k iterations with a small learning rate for best results, these visualisations suggest that faster training for fewer iterations might be sufficient to use for simulation.

\clearpage

\section{Rollout Examples}

\newcommand{\movicrolloutone}{006}
\newcommand{\movicrollouttwo}{005}
\newcommand{\movicrolloutthree}{007}
\newcommand{\movicrolloutfour}{001}

\newcommand{\movibrolloutone}{003}
\newcommand{\movibrollouttwo}{005}

\newcommand{\SDFSIMMOVIC}{SDF-Sim_xid74908215_wid1}

\newcommand{\FIGNOMESHMOVIC}{FIG-no-mesh_xid74908777_wid1}

\newcommand{\SDFSIMMOVIB}{SDF-Sim_xid75391876_wid1}

\newcommand{\FIGNOMESHMOVIB}{FIG-no-mesh_xid75297592_wid1}

\newcommand{\FIGMOVIB}{FIG_xid75178290_wid1}

\newcommand{\MOVICROW}[1]{
    \begin{subfigure}[b]{0.1\textwidth}
    \centering
     Ground\\truth\vspace{1cm}
    \end{subfigure}
    \begin{subfigure}[b]{0.89\textwidth}
    \adjincludegraphics[width=\textwidth,trim={0 0 0 {.5\height}},clip]{figures/rollouts/movic_traj#1_\FIGNOMESHMOVIC.png}
    \end{subfigure}\\
    \begin{subfigure}[b]{0.1\textwidth}
     \centering
     FIGNet*\vspace{1cm}
    \end{subfigure}
    \begin{subfigure}[b]{0.89\textwidth}
    \adjincludegraphics[width=\textwidth,trim={0 {.5\height} 0 0},clip]{figures/rollouts/movic_traj#1_\FIGNOMESHMOVIC.png}
    \end{subfigure}\\
    \begin{subfigure}[b]{0.1\textwidth}
     \centering
     SDF-Sim\vspace{1cm}
    \end{subfigure}
    \begin{subfigure}[b]{0.89\textwidth}
    \adjincludegraphics[width=\textwidth,trim={0 {.5\height} 0 0},clip]{figures/rollouts/movic_traj#1_\SDFSIMMOVIC.png}
    \end{subfigure}\\
}

\newcommand{\MOVIBROW}[1]{
    \begin{subfigure}[b]{0.1\textwidth}
    \centering
     Ground\\truth\vspace{1cm}
    \end{subfigure}
    \begin{subfigure}[b]{0.89\textwidth}
    \adjincludegraphics[width=\textwidth,trim={0 0 0 {.5\height}},clip]{figures/rollouts/movib_traj#1_\FIGNOMESHMOVIB.png}
    \end{subfigure}\\
    \begin{subfigure}[b]{0.1\textwidth}
     \centering
     FIGNet\vspace{1cm}
    \end{subfigure}
    \begin{subfigure}[b]{0.89\textwidth}
    \adjincludegraphics[width=\textwidth,trim={0 {.5\height} 0 0},clip]{figures/rollouts/movib_traj#1_\FIGMOVIB.png}
    \end{subfigure}\\
    \begin{subfigure}[b]{0.1\textwidth}
     \centering
     FIGNet*\vspace{1cm}
    \end{subfigure}
    \begin{subfigure}[b]{0.89\textwidth}
    \adjincludegraphics[width=\textwidth,trim={0 {.5\height} 0 0},clip]{figures/rollouts/movib_traj#1_\FIGNOMESHMOVIB.png}
    \end{subfigure}\\
    \begin{subfigure}[b]{0.1\textwidth}
     \centering
     SDF-Sim\vspace{1cm}
    \end{subfigure}
    \begin{subfigure}[b]{0.89\textwidth}
    \adjincludegraphics[width=\textwidth,trim={0 {.5\height} 0 0},clip]{figures/rollouts/movib_traj#1_\SDFSIMMOVIB.png}
    \end{subfigure}\\
}

\begin{figure*}[hb!]
\centering
\MOVICROW{\movicrolloutone}
\bigskip
\MOVICROW{\movicrollouttwo}
\caption{Rollout comparisons between baselines and SDF-Sim on Kubric Movi-C.}
\label{fig:rollouts}
\end{figure*}

\begin{figure*}
\centering
\MOVICROW{\movicrolloutthree}
\bigskip
\MOVICROW{\movicrolloutfour}
\caption{Rollout comparisons between baselines and SDF-Sim on Kubric Movi-C.}
\label{fig:rollouts2}
\end{figure*}

\begin{figure*}
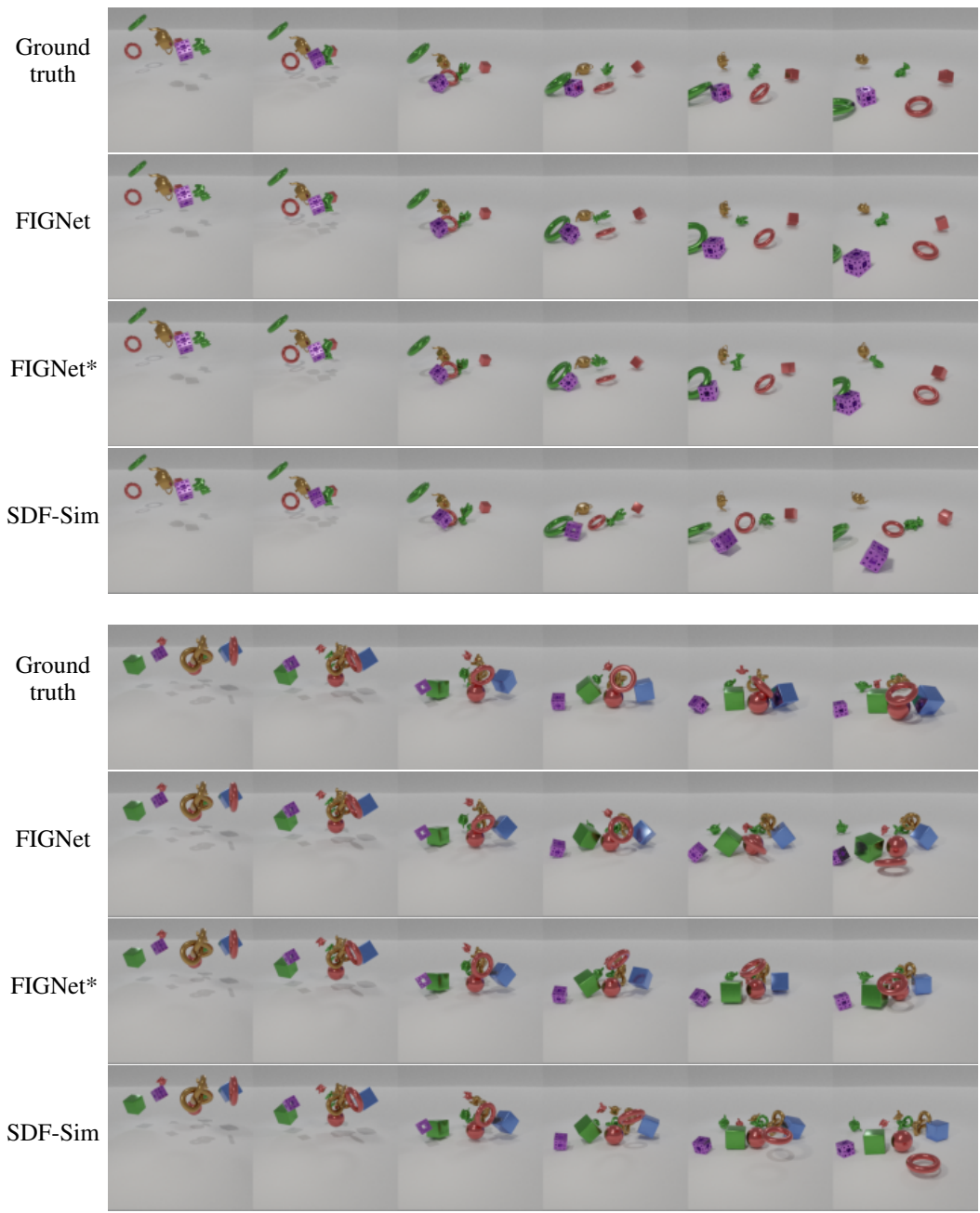

\centering
\MOVIBROW{\movibrolloutone}
\bigskip
\MOVIBROW{\movibrollouttwo}
\caption{Rollout comparisons between baselines and SDF-Sim on Kubric Movi-B.}
\label{fig:rollouts_b}
\end{figure*}

\clearpage

\newpage
\section*{NeurIPS Paper Checklist}

\begin{enumerate}

\item {\bf Claims}
    \item[] Question: Do the main claims made in the abstract and introduction accurately reflect the paper's contributions and scope?
    \item[] Answer: \answerYes{} 
    \item[] Justification: The main claims in the abstract and introduction reflect the scope of the paper and are supported by the experimental results.

\item {\bf Limitations}
    \item[] Question: Does the paper discuss the limitations of the work performed by the authors?
    \item[] Answer: \answerYes{} 
    \item[] Justification: The paper includes an explicit Limitations sections and states the assumptions (e.g. access to 3D meshes). The paper provides detailed investigations on scaling and efficiency.

\item {\bf Theory Assumptions and Proofs}
    \item[] Question: For each theoretical result, does the paper provide the full set of assumptions and a complete (and correct) proof?
    \item[] Answer: \answerNA{} 
    \item[] Justification: The paper does not contain theoretical results.

    \item {\bf Experimental Result Reproducibility}
    \item[] Question: Does the paper fully disclose all the information needed to reproduce the main experimental results of the paper to the extent that it affects the main claims and/or conclusions of the paper (regardless of whether the code and data are provided or not)?
    \item[] Answer: \answerYes{} 
    \item[] Justification: The paper introduces a new model architecture. We provide the detailed description of the architecture, sufficient to reproduce the results.

\item {\bf Open access to data and code}
    \item[] Question: Does the paper provide open access to the data and code, with sufficient instructions to faithfully reproduce the main experimental results, as described in supplemental material?
    \item[] Answer: \answerNo{}
    \item[] Justification: We provide a detailed description of the method sufficient to reproduce it. Open-source release of the code is not possible for this work.

\item {\bf Experimental Setting/Details}
    \item[] Question: Does the paper specify all the training and test details (e.g., data splits, hyperparameters, how they were chosen, type of optimizer, etc.) necessary to understand the results?
    \item[] Answer: \answerYes{} 
    \item[] Justification: We provide the training details, data splits, type of optimizer, etc.

\item {\bf Experiment Statistical Significance}
    \item[] Question: Does the paper report error bars suitably and correctly defined or other appropriate information about the statistical significance of the experiments?
    \item[] Answer: \answerYes{} 
    \item[] Justification: We report the confidence intervals over multiple random seeds for our experiments.

\item {\bf Experiments Compute Resources}
    \item[] Question: For each experiment, does the paper provide sufficient information on the computer resources (type of compute workers, memory, time of execution) needed to reproduce the experiments?
    \item[] Answer: \answerYes{} 
    \item[] Justification: We provide the information about the memory, runtime and hardware for our experiments.
    
\item {\bf Code Of Ethics}
    \item[] Question: Does the research conducted in the paper conform, in every respect, with the NeurIPS Code of Ethics \url{https://neurips.cc/public/EthicsGuidelines}?
    \item[] Answer: \answerYes{}
    \item[] Justification: The authors reviewed the Code of Ethics and confirm that this work follows the code.

\item {\bf Broader Impacts}
    \item[] Question: Does the paper discuss both potential positive societal impacts and negative societal impacts of the work performed?
    \item[] Answer: \answerYes{} 
    \item[] Justification: We provide an overview of the potential societal impact, although we do not foresee our work on physical simulation to have adversarial consequences.
    
\item {\bf Safeguards}
    \item[] Question: Does the paper describe safeguards that have been put in place for responsible release of data or models that have a high risk for misuse (e.g., pretrained language models, image generators, or scraped datasets)?
    \item[] Answer: \answerNA{} 
    \item[] Justification: To our best judgement, the presented approach (scalable 3D simulation) presents no risks for misuse.

\item {\bf Licenses for existing assets}
    \item[] Question: Are the creators or original owners of assets (e.g., code, data, models), used in the paper, properly credited and are the license and terms of use explicitly mentioned and properly respected?
    \item[] Answer: \answerYes{}
    \item[] Justification: Yes, the datasets and the models used as baselines are properly credited and the terms of use respected.

\item {\bf New Assets}
    \item[] Question: Are new assets introduced in the paper well documented and is the documentation provided alongside the assets?
    \item[] Answer: \answerNA{} 
    \item[] Justification: The paper does not release new assets.

\item {\bf Crowdsourcing and Research with Human Subjects}
    \item[] Question: For crowdsourcing experiments and research with human subjects, does the paper include the full text of instructions given to participants and screenshots, if applicable, as well as details about compensation (if any)? 
    \item[] Answer: \answerNA{}
    \item[] Justification: Current paper does not use human participants and does not require an approval.

\item {\bf Institutional Review Board (IRB) Approvals or Equivalent for Research with Human Subjects}
    \item[] Question: Does the paper describe potential risks incurred by study participants, whether such risks were disclosed to the subjects, and whether Institutional Review Board (IRB) approvals (or an equivalent approval/review based on the requirements of your country or institution) were obtained?
    \item[] Answer: \answerNA{}
    \item[] Justification: Current paper does not use human participants and does not require an approval.

\end{enumerate}

\end{document}